\documentclass[crcready,onecolumn]{iosart2x}

\usepackage{amsmath}
\usepackage{rotating}
\usepackage{verbatim}
\usepackage{multicol}
\usepackage{listings}
\usepackage{xcolor}
\newcommand{\ent}[1]{\textsf{#1}}
\newcommand{\ufo}[1]{\textsf{{\normalsize UFO:}#1}}

\newcommand{\brequation}{\notag\\[-6pt]}
\newcommand{\aln}{&}
\newcommand{\bln}{&\quad}

\newcounter{axiomcounter}
\newcounter{axiomcounterb}

\newcommand{\ax}[1]{\stepcounter{axiomcounter}\tag{\textbf{a\theaxiomcounter}}\label{#1}}

\pubyear{0000}
\volume{0}
\firstpage{1}
\lastpage{1}

\usepackage[colorinlistoftodos,textsize=footnotesize]{todonotes}
\usepackage{hyperref}
\hypersetup{
    colorlinks=true,
    linkcolor=blue,
    citecolor=teal,
    filecolor=magenta,
    urlcolor=blue,
    pdftitle={Overleaf Example},
    pdfpagemode=FullScreen,
}

\usepackage{soul} \usepackage{soulutf8} 
\usepackage{amsmath}
\usepackage{enumitem}
\usepackage{xspace}

\usepackage{nameref} 
\makeatletter
\newcommand*{\currentname}{\@currentlabelname}
\makeatother

\newcommand{\ESCONDER}[1]{\VERMELHO{<TEXTO ESCONDIDO>}}

\usepackage{marginnote}
\usepackage{etoolbox}

\newcommand{\ATENCAO}[1]{}

\newcommand{\VERMELHO}[1]{\ATENCAO{VERMELHO} {\color{red}#1}}

\newcommand{\IE}{\emph{i.e.,}\xspace}
\newcommand{\EG}{\emph{e.g.,}\xspace}

\setlist[enumerate,1]{label={(\arabic*)}}

\newcounter{footnotemarknum}
 
\begin{document}

\begin{frontmatter}

\title{Towards an ontology of portions of matter to support multi-scale analysis and provenance tracking}
\runtitle{Towards an ontology of portions of matter}

\begin{aug}
\author[A,B]{\inits{L.V.}\fnms{Lucas Valadares} \snm{Vieira}\ead[label=e1]{lucasvaladares@petrobras.com.br}%
\thanks{Corresponding author. \printead{e1}.}}
\author[B]{\inits{M.}\fnms{Mara} \snm{Abel}\ead[label=e2]{marabel@inf.ufrgs.com.br}}
\author[B]{\inits{F.H.}\fnms{Fabricio Henrique} \snm{Rodrigues}\ead[label=e3]{fabricio.rodrigues@inf.ufrgs.br}}
\author[D]{\inits{T.P.}\fnms{Tiago Prince} \snm{Sales}\ead[label=e4]{t.princesales@utwente.nl}}
\author[D]{\inits{C.M.}\fnms{Claudenir M.} \snm{Fonseca}\ead[label=e5]{c.moraisfonseca@utwente.nl}}
\address[A]{E\&P, \orgname{Petrobras S/A},
Rio de Janeiro, \cny{Brazil}\printead[presep={\\}]{e1}}
\address[B]{Instituto de Informática, \orgname{Universidade Federal do Rio Grande do Sul}, \cny{Brazil}\printead[presep={\\}]{e2,e3}}
\address[D]{Services \& Cybersecurity Group, \orgname{University of Twente}, \cny{The Netherlands}\printead[presep={\\}]{e4,e5}}
\end{aug}

\begin{abstract}
This paper presents an ontology of portions of matter with practical implications across scientific and industrial domains. The ontology is developed under the Unified Foundational Ontology (UFO), which uses the concept of quantity to represent topologically maximally self-connected portions of matter. The proposed ontology introduces the granuleOf parthood relation, holding between objects and portions of matter. It also discusses the constitution of quantities by collections of granules, the representation of sub-portions of matter, and the tracking of matter provenance between quantities using historical relations. Lastly, a case study is presented to demonstrate the use of the portion of matter ontology in the geology domain for an Oil \& Gas industry application. In the case study, we model how to represent the historical relation between an original portion of rock and the sub-portions created during the industrial process. Lastly, future research directions are outlined, including investigating granularity levels and defining a taxonomy of events.
\end{abstract}

\begin{keyword}
\kwd{portion of matter}
\kwd{quantity}
\kwd{mereology}
\kwd{UFO}
\kwd{geology}
\end{keyword}

\end{frontmatter}

\setlength{\abovedisplayskip}{0pt}
\setlength{\abovedisplayshortskip}{0pt}
\setlength{\belowdisplayskip}{12pt}
\setlength{\belowdisplayshortskip}{0pt}

\section{Introduction}\label{s1}

Matter is an essential entity of the physical world. It is often referred to as substance or stuff and is the ultimate constituent of every physical object, ranging from subatomic particles to celestial bodies. Therefore, it is crucial to develop a coherent and comprehensive ontological theory of matter and its portions in order to construct conceptual models that accurately describe the material world.

The significance of an ontology of portions of matter extends beyond theoretical contemplation; it holds practical implications across scientific and industrial domains. Among the possible benefits, two are the focus of this work. First, to reconcile matter and its parts on different scales of observation. And second, to track the provenance of matter, \IE the historical connections between portions of matter.

Matter is typically understood in a single observation level, where it is viewed as a homogeneous entity that constitutes physical objects (such as the steel constituting a bike). However, in fields like Geology and Biomedicine, matter is described across multiple scales of observation. For instance, in the geology domain, there are ontologies concerned with rocks in a coarse scale \citep{abel2015geomodeling}, and others concerned with the discrete parts of a rock \citep{abel2004petrographer}. Likewise, in the biomedical domain, there are ontologies covering macroscopically uniform portions of tissue (\EG eptithelium, neural tissue) or bodily fluids (\EG blood, urine), as well as its microscopic discrete components (\EG cells) \citep{mungall2012uberon, mendonca2013hemocomponents, rodrigues2017urine}.

In such cases, the same matter that appears homogeneous at a given observation scale exhibits a discrete character when analyzed at a finer scale (\textit{e.g.}, a portion of epithelial tissue that seems homogeneous to the naked eye reveals itself as a collection of compactly packed individual cells). To properly deal with this issue, we must resort to certain mereological relationships not yet clearly defined in the ontology literature.

Another issue approached by this work concerns the tracing of the provenance of matter. This is a fundamental requirement for numerous information systems dealing with matter across several knowledge domains. These systems must provide insight into the historical connections between a particular portion of matter that currently constitutes an entity and other portions of matter that existed in the past. For instance, how a rock constituting a volcano is related to lava from a past eruption. Furthermore, they must establish the relationship between two distinct amounts of matter that once composed a single portion of matter but that are currently spatially scattered (\textit{e.g.}, the link between a sample of water collected from a lake for quality inspection and the larger portion of water that remains in the lake).

Lastly, such an ontology can serve as a common language for the community modeling matter entities, reducing semantic overloading, conceptual disagreements, and false agreements. More importantly, it can be the basis for creating informational artifacts that are helpful in applications among the relevant domains.

Given that, this work intends to address the issues mentioned above by answering the following research questions:
\begin{itemize}
    \item RQ1: What is the meronymic relation between a portion of matter and its discrete parts?
    \item RQ2: What constitutes portions of matter?
    \item RQ3: How are portions of matter historically related through their discrete parts?
\end{itemize}

In this paper, we provide an ontological analysis of portions of matter, grounded in the Unified Foundational Ontology (UFO) \citep{guizzardi2022ufo}, using the OntoUML \citep{guizzardi2005} language and further formalizing the necessary axioms using first-order logic. Lastly, we use the theory to a study case involving rocks (a type of portion of matter) in a geology domain context applied to the Oil \& Gas industry.

The remaining of this work is organized as follows: Section \ref{section_background} briefly describes the top-level UFO ontology, which is the basis over which we develop our work; 
Section \ref{section_basic_assumptions} discusses the basic ontological assumptions made; 
Section \ref{section_theory} presents the main contribution of this work, the ontological analysis of portions of matter; 
Section \ref{section_case_study} provides an illustrative case study showing how the proposed ontology and its implementation can aid in a real-world application; 
Section \ref{section_related_work} summarises the related work on the ontological representation of matter in ontology literature and compares to our work; 
Section \ref{section_discussion} discusses some alternative conceptualization and consequences of the ontology proposed; 
and lastly Section \ref{section_conclusion} summarises and concluded the findings in this paper.

\section{Unified Foundational Ontology}
\label{section_background}

The Unified Foundational Ontology (UFO) is a formal reference ontology consisting of a set of micro-theories about fundamental conceptual modeling notions. It has been actively developed and expanded using notions and tools from sources such as other formal ontologies, logic, and cognitive psychology. It is intended to be domain-independent and has been applied to build ontologies in several domains ranging from natural to social sciences \citep[provide a comprehensive list of applications.]{guizzardi2022ufo}.

\subsection{UFO taxonomy}
\begin{figure}
  \centering
  \includegraphics[width=\linewidth]{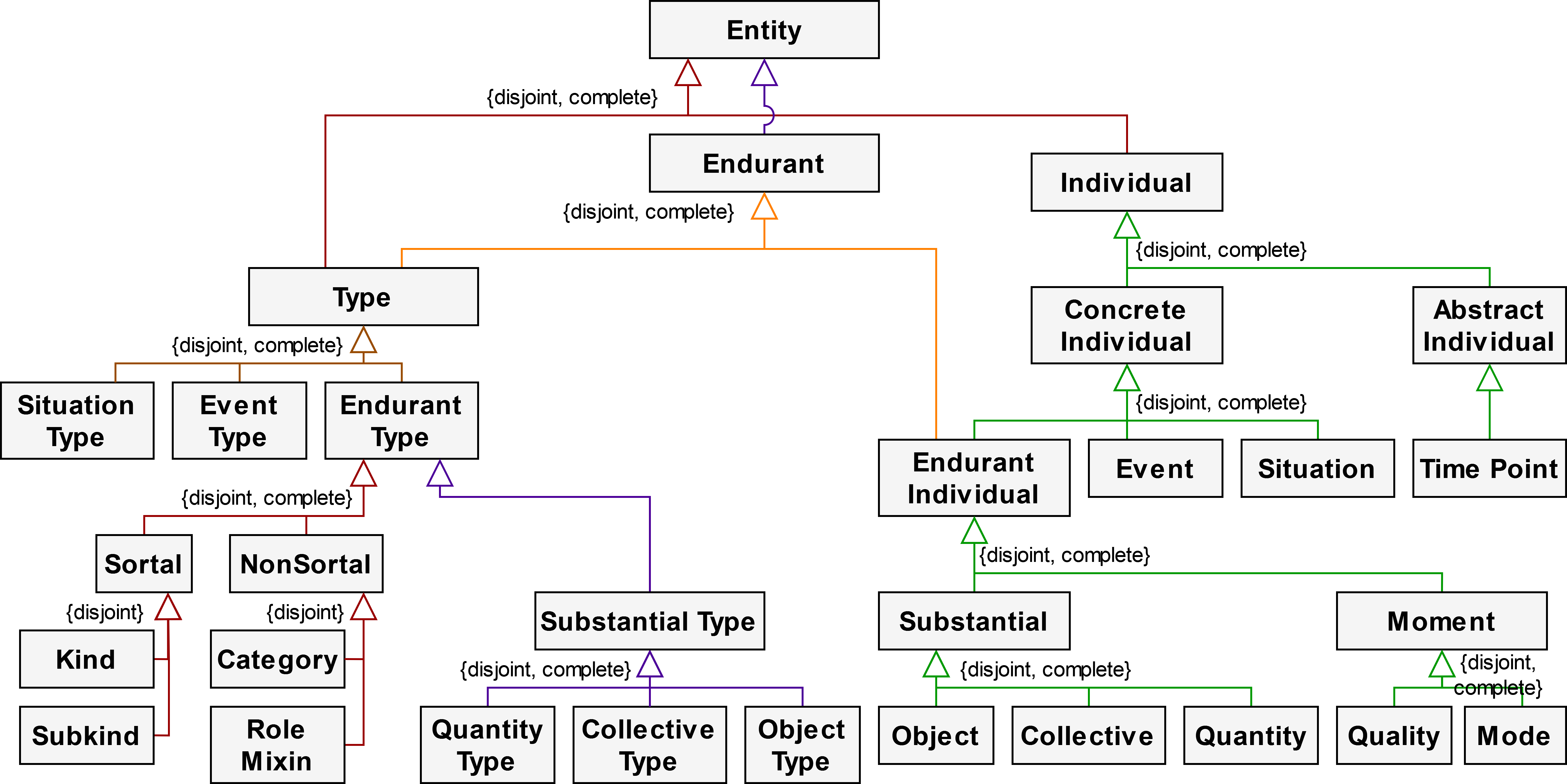}
  \caption{Fragment of the UFO taxonomy, focusing on the top-level entities relevant to this work. Based on \cite{guizzardi2022ufo} and \cite{fonseca2022incorporating}. Colors are only a visual aid to help track distinct partitions of entities.}
  \label{fig_fragment_of_ufo_taxonomy}
\end{figure}

The first division of \textit{things} (\textit{entities}) in UFO (Fig. \ref{fig_fragment_of_ufo_taxonomy}) is between \textit{types} (\textit{universals}) and \textit{individuals} (\textit{particulars}). \textit{Types} are \textit{things} that collect invariant properties exhibited by its instances. \textit{Individuals} are \textit{things} that exhibit properties and can not be instantiated \citep{guizzardi2022ufo, fonseca2022incorporating}.

Another partition of \textit{things} in UFO is between \textit{endurants}, \textit{events} and \textit{situations} \citep{almeida2019events, almeida2019gufo}. \textit{Endurants} are \textit{things} that are completely present whenever they are present and may change over time. \textit{Endurants} are partitioned into \textit{types} and \textit{endurant individuals} \citep{fonseca2022incorporating}. \textit{Events} are \textit{individuals} that unfold in time accumulating temporal parts \citep{guizzardi2022ufo}. \textit{Situations} are \textit{individuals} that describe particular configurations of parts of the world that can be understood as a whole \citep{guizzardi2013events}.

\textit{Types} are further partitioned into \textit{situation types}, \textit{event types} and \textit{endurant types}, if their instances are, respectively, \textit{situations}, \textit{events} or \textit{endurants}. \textit{Endurant types} are classified into two distinct partitions: the first one is based on the nature of its instances, and the second partition is based on meta-properties (rigidity, sortality, and identity principle). The first partition of \textit{Endurant Type} is between \textit{Moment Type} and \textit{Substantial Type}, and the latter is partitioned into \textit{Quantity Type}, \textit{Collective Type} and \textit{Object Type}.

The second partition of \textit{Endurant Types} includes \textit{rigid types}, meaning that they capture patterns of essential properties (not contingent) of their instances. A \textit{Rigid type} is a \textit{category}, if it is also a \textit{Non-Sortal Type} (instances have distinct identity criteria); a \textit{Kind}, if it is a \textit{Sortal Type} (instances with single identity criteria) and provides identity criteria to its instances; or a \textit{Subkind}, if it is a \textit{Sortal Type} and inherits the identity criteria from a \textit{Kind}. \textit{Endurant Types} can also be \textit{AntiRigid Types}, based on contingent properties of its instances. \textit{Phase Mixin} and \textit{Role Mixin} are \textit{AntiRigid Types} that are also \textit{NonSortal Types}, the former captures intrinsic contingent properties of its instances, and the latter captures relational properties.

\textit{Endurant individuals} are either \textit{substantials} (existentially independent entities) or \textit{moments} (existentially dependent entities). \textit{Substantials} are partitioned, according to their unity conditions, in \textit{objects} (functional complexes), \textit{collectives}, and \textit{quantities}.

\subsection{Matter entities in UFO}

UFO defines two concepts to model matter entities. First, \textit{amounts of matter} are \textit{substantials} with no unity \citep{guizzardi2005}, and are mereologically invariant. As they do not have unity, \textit{amounts of matter} do not have a counting principle and are homeomerous. And, since they are mereologically invariant, all of their parts are essential.

\textit{Quantity} is the second concept to represent matter entities in UFO \citep{guizzardi2005, guizzardi2010}. \textit{Quantities} are maximally-topologically-self-connected portions of amounts of matter. As \textit{amounts of matter}, they are also mereologically invariant \citep{guizzardi2010}. Differently from \textit{amounts of matter}, \textit{Quantities} are necessarily maximal, meaning that no \textit{quantity} is part of a \textit{quantity} of the same kind \citep{guizzardi2010}, and consequently, they are not homeomerous.

Although the concept of \textit{amount of matter} is included in UFO's ontological foundation \cite{guizzardi2005}, \textit{quantity} is the recommended choice to represent matter entities in UFO \cite{guizzardi2010, guizzardi2022ufo}. \textit{Quantity} is part of OntoUML \cite{guizzardi2005} and gUFO \cite{almeida2019gufo} implementations, and it will be used in this work to represent \textit{portions of matter}.

Regarding mereological relations between \textit{quantities}, \cite{guizzardi2010} proposes the \textit{subQuantityOf} relation. \textit{subQuantityOf} is a proper parthood relation that holds between two quantities of different kinds \cite{guizzardi2010}. For example, alcohol is a \textit{subQuantityOf} of wine.

\subsection{Properties of part-whole relations} \label{subsection_prop_part_whole}
\cite{guizzardi2005} introduce the following important concepts to characterize part-whole relations. First, they introduce \textit{existential dependence}, which states that an entity \textit{x} is \textit{existentially dependent} on an entity \textit{y} if it is necessary that \textit{y} exists for \textit{x} to exist. They also introduce \textit{generic dependence}, according to which an entity \textit{x} is generically dependent of a type \textit{T} if it is necessary an instance of \textit{T} for \textit{x} to exist.

An entity \textit{x} is an \textit{essential part} of another entity \textit{y} if it is necessary that at every time that \textit{y} exists, \textit{x} is a proper part of \textit{y}. For example, the brain is an essential part of a person for a particular person to exist, its brain has to exist. An entity \textit{x} is a \textit{mandatory part} of entity \textit{y} if \textit{y} is generically dependent of type \textit{T} which \textit{x} is an instance. For instance, the heart is a mandatory part of a person since heart transplants are possible.

An entity \textit{x} is a \textit{inseparable part} of an entity \textit{y} if x is existentially dependent and necessarily a part of \textit{y}. An entity \textit{y} is a mandatory whole of an entity \textit{x} if \textit{x} is \textit{generically dependent} on the type \textit{T} that \textit{y} instantiates, and \textit{x} is necessarily part of an instance of \textit{T}.

\section{Basic Assumptions}\label{section_basic_assumptions}

Following UFO, we assume the \textit{three-dimensionalism} view on concrete objects \citep{maureen2016threedimensionalism, Emery2020time}, which means that they are wholly present at any time point in which they are present. Therefore, endurants do not have temporal parts. Additionally, we adopt the \textit{growing block view} \citep{Emery2020time} according to which both present and past entities exist in each possible world. Under this view, the future does not exist, and each new instant monotonically expands the set of possible worlds.

Regarding the existence of entities, UFO \citep{guizzardi2022ufo} introduces an \textit{existence} predicate, which is defined for any possible entity. Adding to the formalization proposed in UFO, we understand that existence in a particular world depends on whether the entity was created by some event. Termination events, however, do not end the existence of an entity but simply modify its ontological status \citep{almeida2019events}. 

To avoid ambiguity, we adhere to the following definitions of terms throughout the remainder of this work. \textit{Stuff} will be used to refer to matter that is formed by atoms (in the chemical sense), equivalently to the way the term is used in \citep{keet2014core}. This statement implies that there is matter that is not stuff, \EG sub-atomic particles (such as protons, electrons, and quarks) are constituted of matter that is not stuff. We will use \textit{amount of matter} to refer to a particular (individual) instance of \textit{stuff} \citep{guizzardi2005}. The term \textit{portion of matter} is equivalent to \ufo{Quantity}, and they will be used interchangeably.

Lastly, we need to discuss the notion of \textit{homeomerosity} due to its use as a property of \textit{amounts of matter} in UFO \citep{guizzardi2010, guizzardi2022ufo}. An \textit{homeomerous} entity is composed of proper parts that are the same kind as itself. For instance, we can say that the amount of water in a cup has two amounts of water as parts, one in the lower half of the cup and another in the upper half. If all instances of the amount of water are homeomerous, each water part has other smaller water parts, leading to infinite regress \citep{guizzardi2010}. We define a type as \textit{infinitely homeomerous} if all its instances are \textit{homeomerous}.

Another possible view is that some types of amount of matter are \textit{finitely homeomerous}, which means that some of their instances are \textit{homeomerous}, but there is a minimal amount of that type of matter that is \textit{non-homeomerous}. Using the example of an amount of water again, it is reasonable to say that if an amount of matter \textit{w} is composed of just a couple of $H_2O$ molecules, then there is no smaller amount of water that can be a proper part of \textit{w}. Hence, that particular minimal amount of water is \textit{non-homeomerous}.

Quantities in UFO are, by definition, \textit{non-homeomerous}. As they are maximally-topologically-self-connected portions of matter, no proper part of a quantity \textit{q} can be a quantity of the same type of \textit{q} since all of such parts will be topologically connected to the others. To illustrate, let us bring back our cup of water once more. The portion of water \textit{w} inside the cup indeed has its upper and lower halves as proper parts, but such parts are not \textit{portions} of water themselves since they are not maximally-topologically-self-connected.

Now, if we pour half of \textit{w} into another cup, then we will have two new maximally-topologically-self-connected portions of water (\IE two quantities) \textit{w\textsubscript{1}} and \textit{w\textsubscript{2}}, but they will not be proper parts of the original portion \textit{w}. This is the case because the portion \textit{w} terminates at the exact moment it is divided into two portions. Consequently, there could be no mereological relation between \textit{w} and \textit{w\textsubscript{1}} or \textit{w\textsubscript{2}}.

For the remainder of this work, we will assume that the amounts of matter are \textit{finitely homeomerous}. Consequently, as UFO's quantities are portions of amounts of matter, this means that, in our view, they are composed of objects. In the next section, we will detail and formalize the implications of this ontological decision.

\section{A complementary analysis of portions of matter}\label{section_theory}

In this section, we present an ontological analysis of portions of matter, complementing previous work about quantities in UFO \citep{guizzardi2005, guizzardi2010}. First, we define a new type of meronymic relation called \ent{granuleOf}, which holds between objects that are parts of a quantity. Second, we discuss the material constitution of quantities. Third, we define historical relations to track the provenance of granules and the genidentity of quantities. Lastly, we propose a partition of \ufo{Quantity} based on its origins.

To support our exposition, we provide diagrams in OntoUML conceptual modeling language \citep{guizzardi2005}, whose built-in restrictions implement UFO formal theory. OntoUML includes a set of modeling primitives, represented as stereotypes in UML class diagrams, that refer to leaves of the UFO taxonomy (Fig. \ref{fig_fragment_of_ufo_taxonomy}).

\subsection{Granules and their relation to quantities}\label{section_granule_parthood}

\begin{figure}
  \centering
  \includegraphics[width=0.7\linewidth]{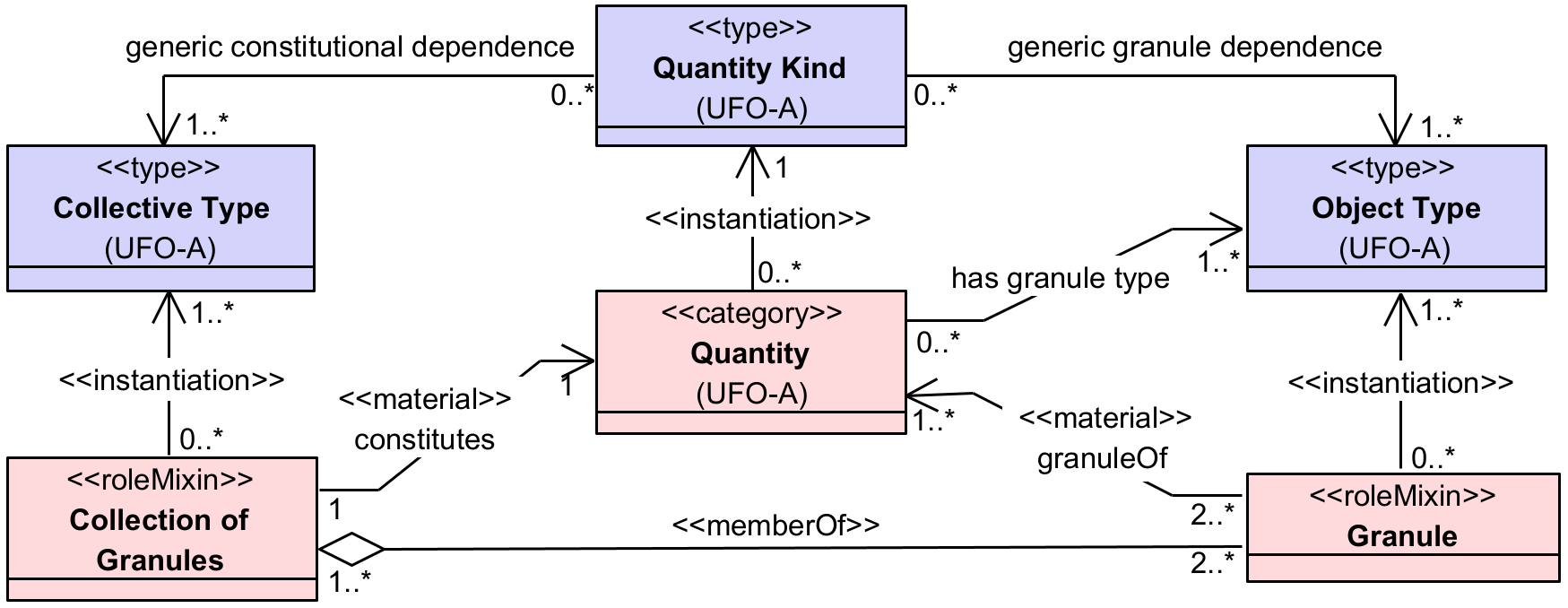}
  \caption{OntoUML diagram showing the relation between granules, quantities, Object Types and Quantity Kinds.}
  \label{fig_relations_granule_quantity}
\end{figure}

In section \ref{section_basic_assumptions}, we introduced the idea that instances of \textit{stuff} are ultimately composed of discrete parts. Following this assumption, here we propose the \ent{granuleOf} meronymic relation, holding between an \ufo{Object} part and its \ufo{Quantity} whole at some \ufo{Time Point} (\ref{a1}). Additionally, we propose the \ent{Granule} type, which is instantiated by every \ufo{Object} that is in a \ent{granuleOf} relation. Considering that being a granule is a contingent property of objects with different identity principles, \ent{Granule} is an anti-rigid non-sortal property, \IE an instance of \ufo{RoleMixin} and \ufo{ObjectType} (Fig. \ref{fig_relations_granule_quantity}).

\begin{figure}
  \centering
  \includegraphics[width=0.7\linewidth]{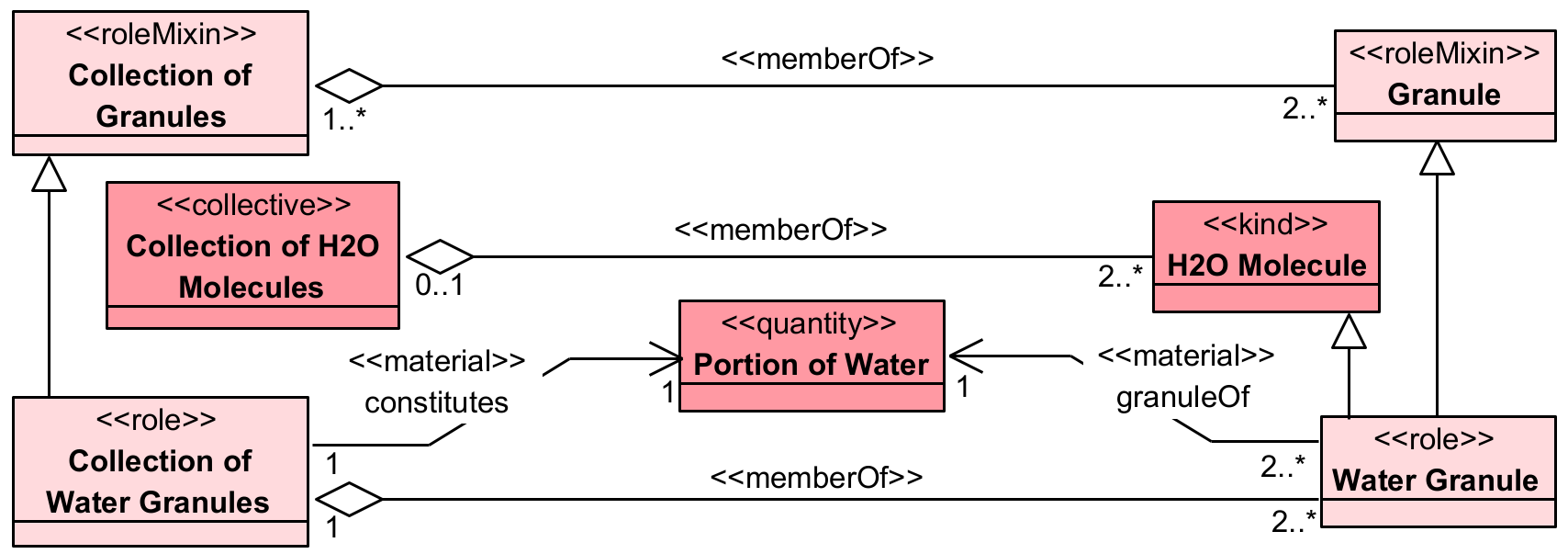}
  \caption{The water constitution and granules pattern.}
  \label{fig_water_constitution_granules}
\end{figure}

The example of water constitution and granular composition (Fig. \ref{fig_water_constitution_granules}) illustrates the relations between those entities. Consider \ent{Portion of Water} as a specialization of \ufo{Quantity} and an instance of \ufo{Kind}, and \ent{H\textsubscript{2}O Molecule} as a specialization of \ufo{Object} and an instance of \ufo{Kind}. Every instance of \ent{Portion of Water} is related to at least two \ent{H\textsubscript{2}O Molecule} through the \ent{granuleOf} relation. While holding this relation, an individual \ent{H\textsubscript{2}O Molecule} instantiates \ent{Water Granule} type, which specializes \ent{Granule} and is an instance of \ufo{Role}.

Furthermore, the \ent{granuleOf} relation follows extensional mereology with strong supplementation. Consequently, every \ufo{Quantity} has at least two distinct instances of \ufo{Object} as its granules. A consequence of strong supplementation is that if a \ufo{Quantity} \textit{q'} has a sub-quantity \textit{q''}, then the granules of \textit{q''} are also granules of \textit{q'} (\ref{a2}). For instance, a portion of wine has a portion of alcohol as a sub-quantity, therefore, the granules of the portion of alcohol are also granules of the portion of wine.

\begin{align}
    \aln \ent{granuleOf}(o, q, t) \rightarrow \ufo{Object}(o) \wedge \ufo{Quantity}(q) \wedge \ufo{TimePoint}(t) \wedge \ent{PP}(o, q, t) \ax{a1} \\
    \aln \ent{subQuantityOf}(q'', q') \rightarrow \forall o(\ent{granuleOf}(o, q'') \rightarrow \ent{granuleOf}(o, q')) \ax{a2} 
\end{align}

Regarding mereotopology, since quantities are defined as topologically self-connected entities, each \textit{Granule} of a \ufo{Quantity} is externally connected to at least another \ent{Granule} of the same \ufo{Quantity}. This means that the granules share at least a point in the border of their spatial region. And every \ent{Granule} of the same \ufo{Quantity} is topologically connected to every other \ent{Granule}, meaning that there is a continuous spatial path between them.

According to the taxonomy of parts proposed by \cite{galton2018yet}, a \ent{Granule} can be classified as an \ent{attached independent part} of a \ufo{Quantity}. This type of parthood is instantiated if the parts are physically connected to at least another part of the whole. Lastly, the \ent{granuleOf} relation is intransitive, non-homeomerous and non-functional \citep[sensu][]{winston1987taxonomy}.

For some types of quantity, especially those more complex ones, formed by different types of granules, the classification as non-functional is debatable. In these quantities (\EG wine), there seem to be different functional roles played by different granules (\EG distinct types of molecules in a portion of wine are related to different properties of the whole). However, this subject requires further research to check whether other relations or entities exist in such cases. For now, we assume that instances of \ufo{Quantity} emerge from an arguably homogeneous relation between their granules.

The \ent{granuleOf} relation captures a more profound relationship between the object and the quantity than a mere spatial parthood. Many moments specific to the type of quantity are dependent existentially on moments of the granule and the relationship among them. For instance, a portion of water has a specific melting, boiling temperature, and solvent capacity related to the properties of its H\textsubscript{2}O granules and their relationships (polarity and hydrogen bonds). However, although the H atoms are components of the granules and, therefore, parts of the water, they are not granules of the water. Consequently, we attribute the \ent{granuleOf} relation as intransitive.

According to the secondary properties of a part-whole relation (\ref{subsection_prop_part_whole}) a \ent{Granule} is an \textit{essential part} of a \ent{Quantity}. Moreover, there is a \textit{generic dependence} \citep[sensu][]{guizzardi2005} between quantities and some \ufo{Object Type} (see Section \ref{subsection_prop_part_whole}). For example, if a portion of water exists, then there must exist some H\textsubscript{2}O granules that are its parts.

We define the \ent{hasGranuleType} relation holding between a \ufo{Quantity} and an \ufo{Object Kind} (Fig. \ref{fig_relations_granule_quantity}) as a specialized generical dependence relation. It means that at least one individual of that object type is a granule of that individual quantity. It is worth noting that quantities are usually made of large amounts of granules and, for most applications, it is unnecessary to keep track of their individual granules. The \ent{hasGranuleType} relation is useful in this case, allowing us to abstract out of the model the individual granules that compose a quantity while keeping some information about the types of such granules.

Lastly, there is a \ent{generic granular dependence} relation (\ent{ggd}) holding between some \ent{Quantity Kind} and some \ent{Object Type} (Fig. \ref{fig_relations_granule_quantity}), as instances of any \ent{Quantity Kind} are characterized by having specific types of object as their granules. For instance, \ent{Water} as a \ent{Quantity Kind} has a generical granular dependence of \ent{H\textsubscript{2}O Molecule}, as for every instance of \ent{Water} there must be a \ent{H\textsubscript{2}O Molecule} as its granule.

\begin{align}
    \aln \ent{ggd}(q', o') \rightarrow \ufo{QuantityKind}(q') \wedge \ufo{ObjectType}(o') \wedge \brequation
    \bln \forall q, t(\ent{iof}\footnotemark{}(q, q', t) \rightarrow \exists o(\ent{iof}(o, o', t) \wedge \ent{granuleOf}(o, q, t))) \ax{aa1}\\
    \aln\ent{hasGranuleType}(q, o') \rightarrow \ufo{Quantity}(q) \wedge \ufo{ObjectKind}(o') \wedge\brequation
    \bln\exists o, t(\ent{iof}(o, o', t)  \wedge \ent{granuleOf}(o, q, t)) \ax{a8}
\end{align}

\footnotetext{We use \ent{iof} as short for \ent{instance of}.}

\subsection{Constitution of quantities}
Using the notion of \textit{material constitution} \citep{baker2000persons}, we define quantities as being constituted by the collection of its granules. This is similar to the conceptualization proposed by \cite{garcia2019rocks} specifically for rocks. In this work, we propose that this notion is true for quantities in general. Therefore, a \ent{Collection of Granules} is a specialization of \ufo{Collective} and an instance of \ufo{RoleMixin}, that aggregates instances of \ent{Collective} which are constituting some \ufo{Quantity} (Fig. \ref{fig_relations_granule_quantity}).

Even though quantities are constituted by collections, not all kinds of collections can constitute quantities. The adequate kind of \ufo{Collective} that can possibly constitute some kind of \ufo{Quantity} must be defined by domain ontologies. Regardless, for a \ufo{Collective} to constitute a \ufo{Quantity}, it must  be in some favorable conditions, and its members being topologically connected is one of these conditions.

\begin{figure}
    \centering
    \includegraphics[width=0.8\linewidth]{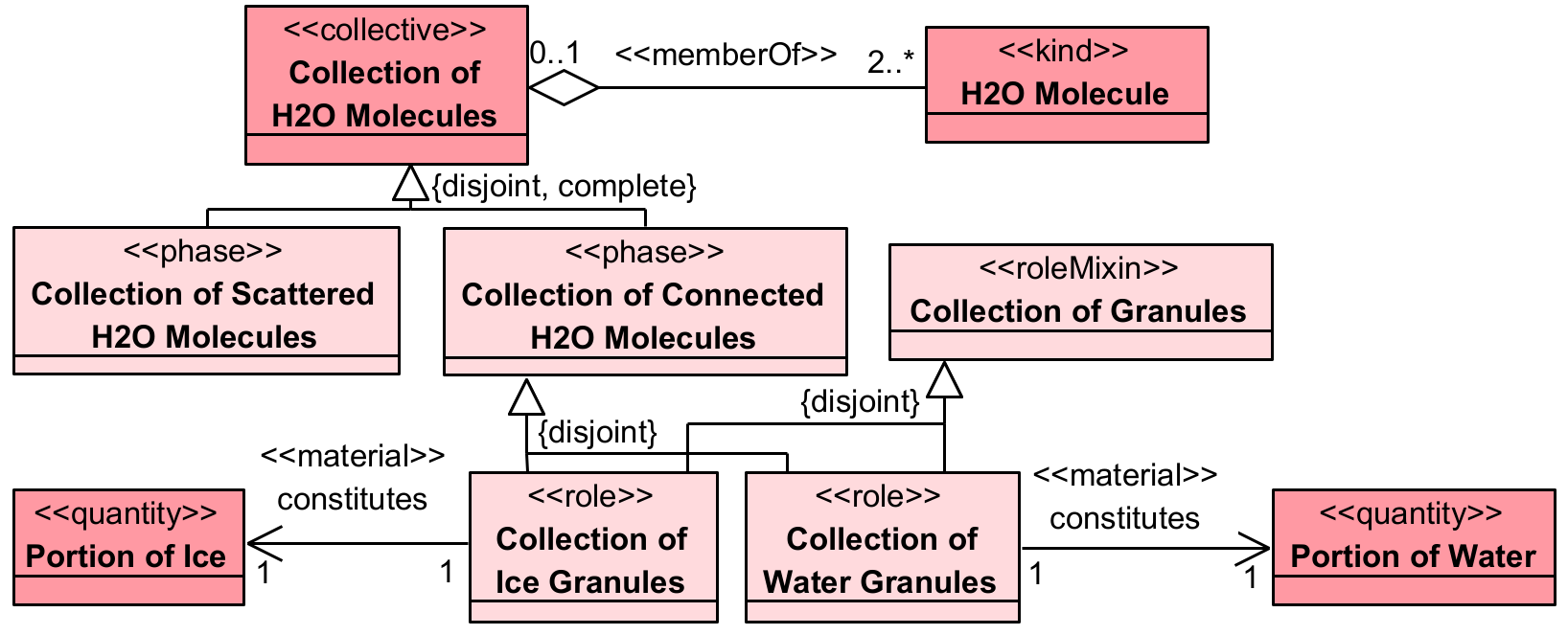}
    \caption{OntoUML diagram with the model of Collection of H\textsubscript{2}O and its anti-rigid specializations.}
    \label{fig_collection_phases_and_roles}
\end{figure}

The models in Fig. \ref{fig_water_constitution_granules} and Fig. \ref{fig_collection_phases_and_roles} illustrate this conceptualization. In those models, the instances of \ent{Collection of H2O Molecules}, which have instances of \ent{H2O Molecule} as members, are partitioned in the \ent{Collection of Scattered H2O Molecules} and \ent{Collection of Connected H2O Molecules} phases. Collections in the latter phase can assume the role of \ent{Collection of Granules}. Specifically, a \ent{Collection of Connected H2O Molecules} can be a \ent{Collection of Water Granules} if constituting a \ent{Portion of Water} or a \ent{Collection of Ice Granules} if constituting a \ent{Portion of Ice}.

\subsection{Granule inheritance}\label{section_granule_inheritance}

By definition, instances of \ufo{Quantity} are \textit{extensional individuals} \citep{guizzardi2010}, which means that all their parts are \textit{essential parts}. Consequently, if a previously existent \ent{granule of} relation between an \ufo{Object} and a \ufo{Quantity} ceases to exist, that particular quantity is terminated by some \ufo{Event} (\ref{h1}). However, an \ufo{Object} is a \textit{separable part} of the \ent{Quantity} whole. This means that an \ufo{Object} might survive the termination of the \ufo{Quantity} whole and, consequently, can be part of different quantities at different times.

\begin{align}
    \aln\ent{granuleOf}(o, q, t') \wedge \exists t''(\neg\ent{granuleOf}(o, q, t'') \wedge t''>t') \rightarrow \exists e(\ufo{Event}(e) \wedge \brequation
    \bln\ent{terminatedBy}(q, e))\ax{h1}
\end{align}

An important part of tracking the provenance of \textit{portions of matter} is determining other quantities from which they inherit their granules. To represent this notion, we define the historical relation \ent{inheritedGranulesFrom} that holds between two quantities that participate in a \ent{Granule Transfer} event.

\begin{figure}
    \centering
    \includegraphics[width=0.8\linewidth]{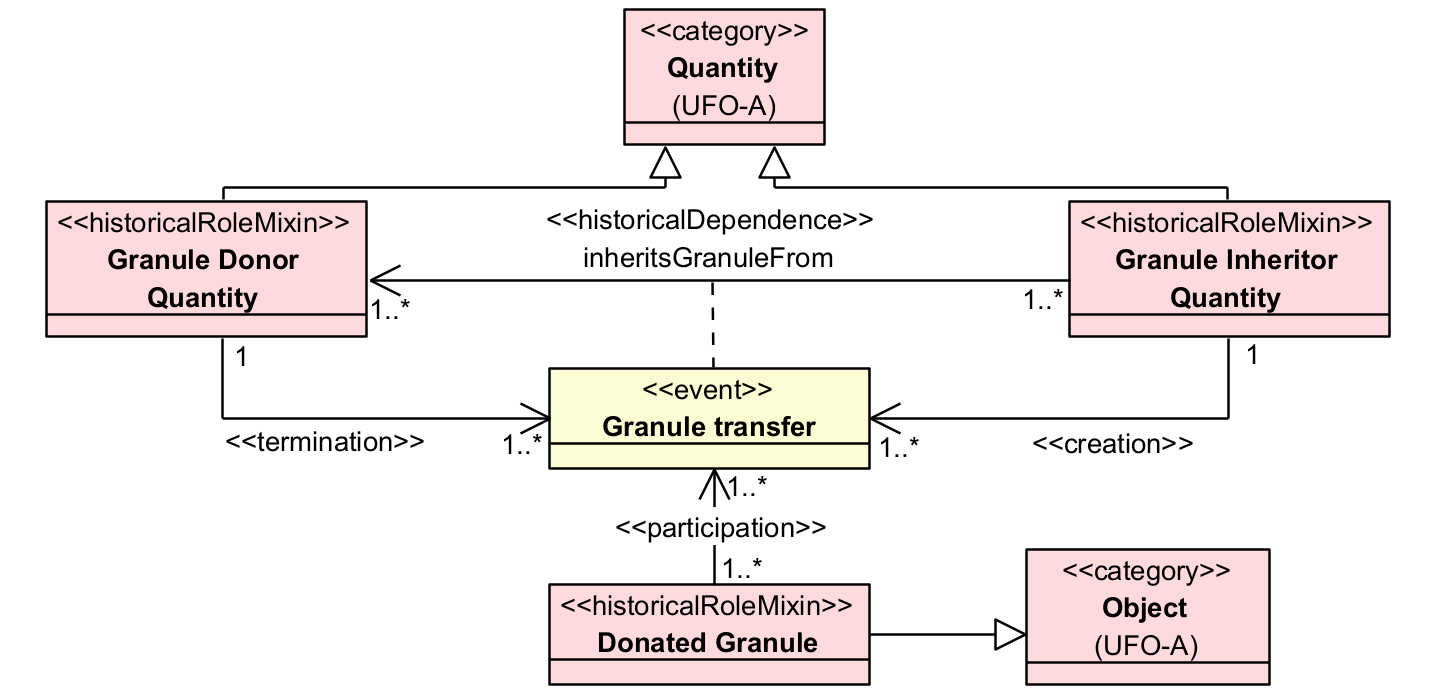}
    \caption{Diagram of the \ent{Granule Transfer} event, that occurs when a \ent{Inheritor Quantity} receives granules from a \ent{Donor Quantity}.}
    \label{fig_granule_inheritance}
\end{figure}

A \ent{Granule Transfer} is an \ufo{Event}, in which some granules from a \ent{Donor Quantity} are transferred to a \ent{Inheritor Quantity}. The granules participating in such an event instantiate \ent{Donated Granule}. Moreover, the \ent{Donor Quantity} is terminated, and the \ent{Inheritor Quantity} is created during the \ent{Granule Transfer} event. Moreover, since \ent{Donor Quantity}, \ent{Inheritor Quantity}, and \ent{Donated Granule} are non-sortal types that entities instantiate in virtue of participating in an event, such types are classified as \textit{historicalRoleMixin} (Fig. \ref{fig_granule_inheritance}).

As stated previously, we have the historical relation \ent{inheritedGranulesFrom} between a \ent{Inheritor Quantity} and a \ent{Donor Quantity}. Following the idea that historical relations have their \textit{truthmakers} in events \cite{fonseca2019relations}, the \ent{inheritedGranulesFrom} have \ent{Granule Transfer} as its \textit{truthmaker}.

Furthermore, the \ent{inheritedGranulesFrom} relation is transitive, as quantities can inherit granules through sequences of \ent{Granule Transfer} events. The \ent{inheritedGranulesFrom} relation is also asymmetric, so we define the \ent{donatedGranulesTo} relation as its inverse.

A quantity \textit{x} can completely inherit its granules from some other quantity \textit{y}, if all the grains in \textit{x} came in \textit{x}. And a quantity \textit{x} can partially inherit its granules from another quantity \textit{y} if some, but not all, granules in \textit{x} come from \textit{y}. Likewise, granule donation can be complete or partial.

\subsection{Sub-portion of stuff}\label{section_subportion}
As UFO's quantities are maximally self-connected \citep{guizzardi2010}, at any single time point, there cannot be any parts of a quantity that are of the same kind. For example, using quantities, there is no way to represent the top and bottom halves of the \textit{portion of wine} in a glass. Suppose some \textit{portion of wine} is split into two glasses, creating two new instances of \textit{portion of wine} and terminating the original portion. In such case, we can use the \ent{inheritedGranulesFrom} to track the historical relation between the newly created portions of wine and the single portion that existed before in the glass. However, in cases like this, not only does each of the new quantities inherit all its granules from the same donor quantity, but they are also of the same kind of donor quantity and, naturally, can be expected to have properties very similar to such a quantity. Even so, we cannot say that the new portions of wine are part of the original portion since it was destroyed when split.

To represent this specific historical relation, we define the \ent{subPortionOf} relation as a specialization of the \ent{inheritedGranuleFrom}. This is a relation linking a \ent{Donor Quantity} and an \ent{Inheritor Quantity} of the same \ufo{Quantity Kind}, such that the collection of granules constituting the inheritor quantity is a subset of the collection of granules that used to constitute the donor quantity.

Accordingly, we define a \ent{Sub-Portion} as a \ufo{Quantity} that holds a \ent{subPortionOf} relation with another \ufo{Quantity}. Additionally, we define an \ent{Original Portion} as a \ufo{Quantity} that is not a sub-portion of another \ufo{Quantity}. Both \ent{Original Portion} and \ent{Sub-portion} are specializations of \ufo{Quantity}, forming a complete and disjoint set, and are rigid non-sortal types that instantiate \ufo{Category}.

A consequence of this disjoint and complete partition is that every instance of \ufo{Quantity} is either an \ent{Original Portion} or a \ent{Sub-Portion}. This is intended to distinguish quantities according to their origins. Some instances of portions of matter resulted from ``reductions'' of a larger portion of the same type, while others are created by other types of events.

Furthermore, the \ent{Sub-Portion Formation} is a specialization of \ent{Granule Transfer} that is the \textit{truthmaker} of the \ent{subPortionOf} relation. In a \ent{Sub-Portion Formation}, the \ent{Granule Inheritor Quantity} is necessarily a \ent{Sub-Portion}.

The \ent{SubPortionOf} relation is transitive, asymmetric, and irreflexive. According to the taxonomy proposed in \citep{galton2018yet}, the \ent{subPortionOf} relation falls under \ent{detached extrinsic part}, as it is about an independent entity (the smaller portion) that was created by being separated from its ``former whole''.

\begin{figure}
    \centering
    \includegraphics[width=0.7\linewidth]{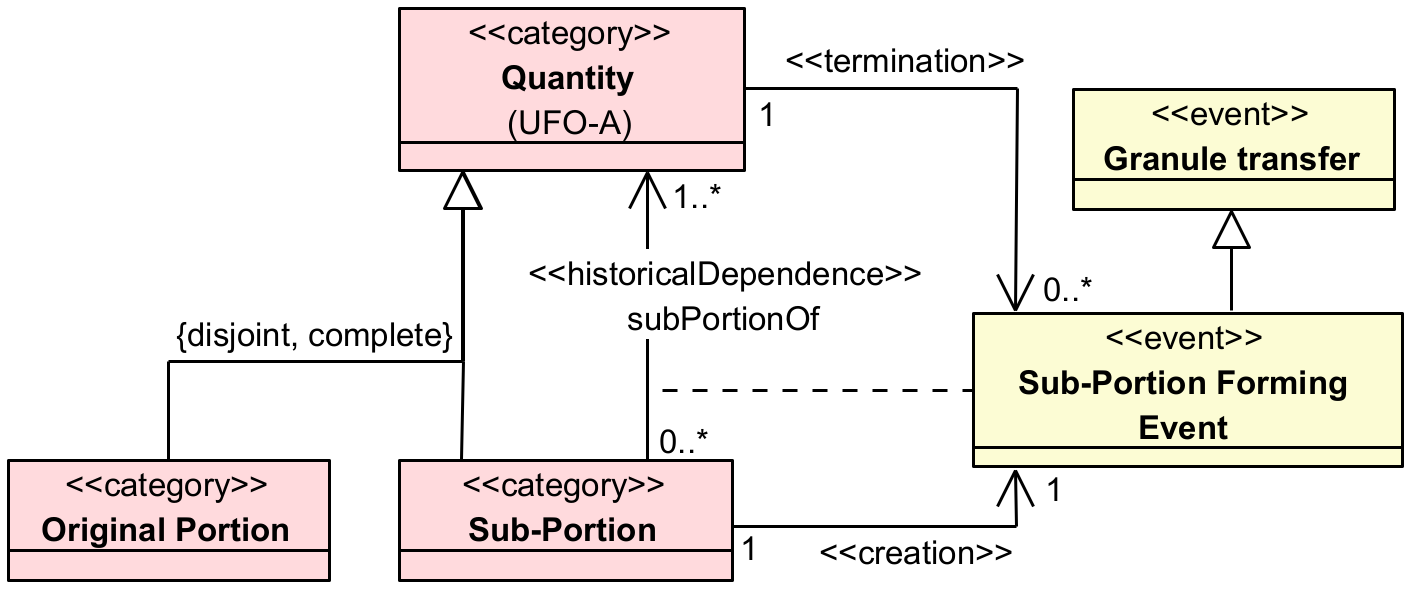}
    \caption{Diagram of the \ent{Granule Transfer} event, that occurs when a \ent{Inheritor Quantity} receives granules from a \ent{Donor Quantity}.}
    \label{fig_subportion}
\end{figure}

\section{Illustrative Case Study}\label{section_case_study}

This section presents an illustrative case study in which we use the ontology proposed in the previous section to model domain knowledge in Geology. Our main objective is to show how ontology-based conceptual models can be used to solve issues in systems that need to track \textit{portions of matter} through a series of events and to describe their relations to other entities on different levels of granularity. The case does not deal with real-world data, but it covers entities and issues similar to those found in industrial information systems.

\begin{figure}
  \centering
  \includegraphics[width=\linewidth]{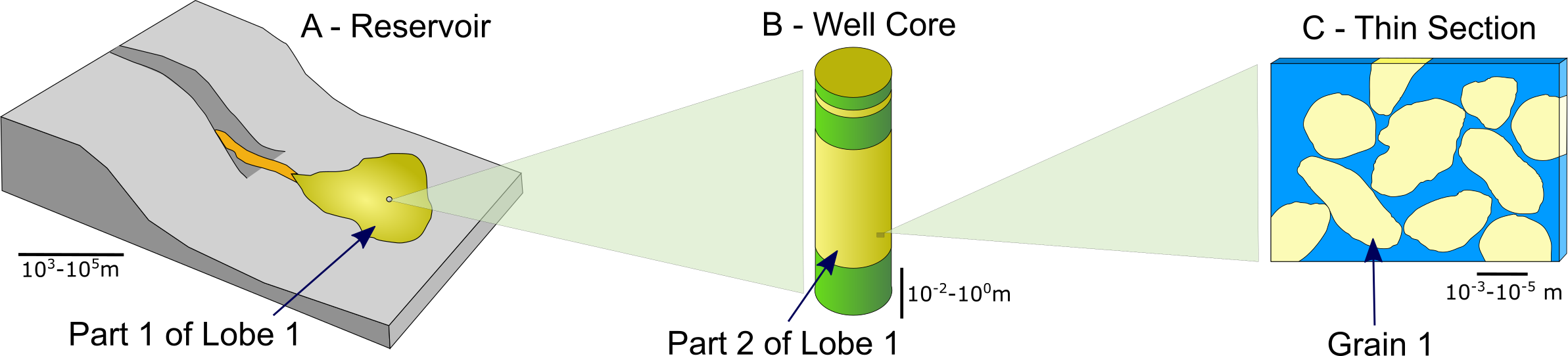}
  \caption{The development of an oil field requires different scales of analysis of the same reality. The most important are the reservoir, core, and microscopic scales of study.}
  \label{fig_app_1}
\end{figure}

Particularly, the case primarily concerns portions of rock and their relation with natural objects and the industrial artifacts created or used during an oil field production in an Oil\&Gas company. In this setting, it is common to model entities observed in three different spatial scales (Fig. \ref{fig_app_1}). Each of these scales focuses on entities of distinct types, with different specialists working on them, and with data obtained from these entities being stored in distinct tables in the company's databases.

A crucial challenge when designing the conceptual models that will be used to implement the information systems is establishing suitable links between the entities represented in such tables. Thus, we will show how the proposed ontology can aid in achieving this goal.

\begin{figure}
  \centering
  \includegraphics[width=\linewidth]{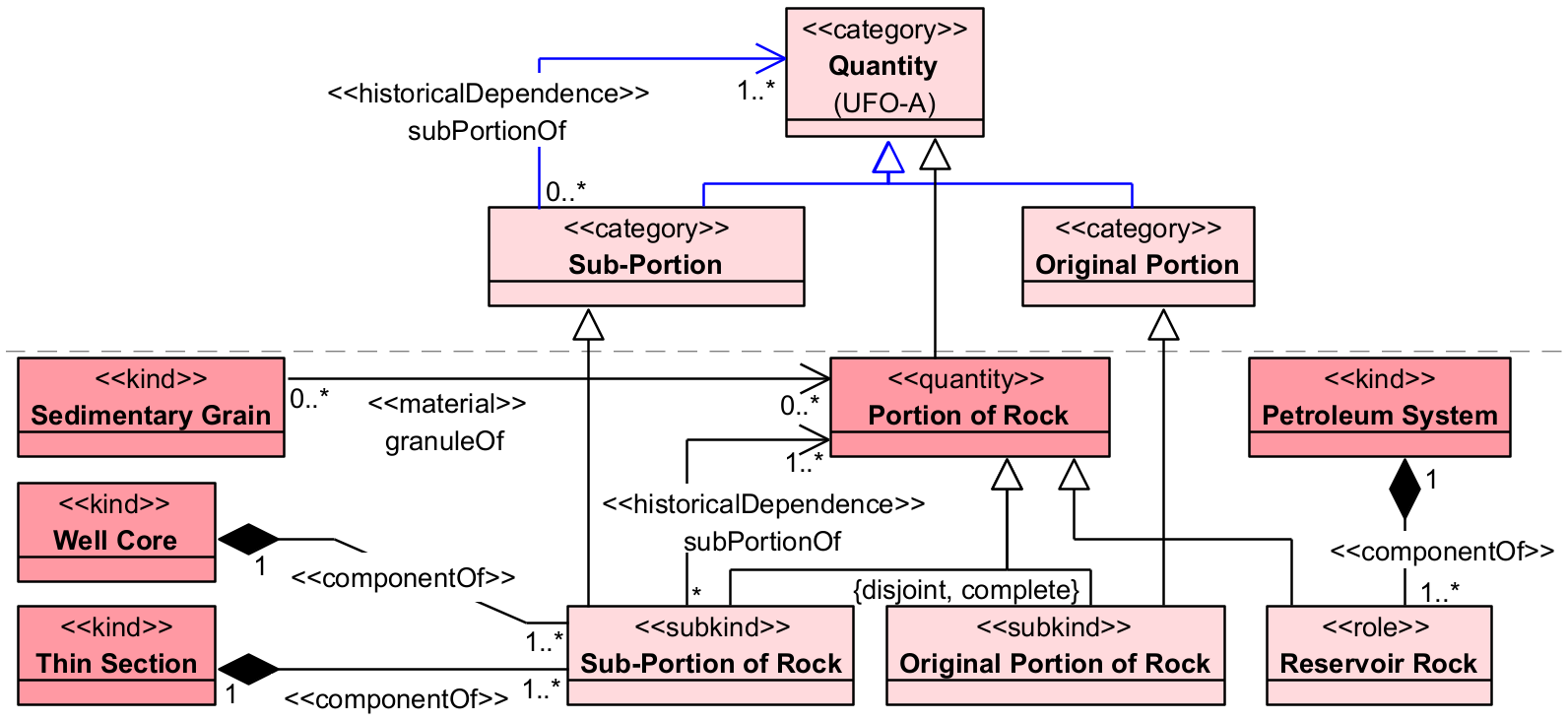}
  \caption{Class diagram showing the how \ent{Portion of Rock} is related to other entities in the Oil\&Gas domain.}
  \label{fig_app_types}
\end{figure}

The first and bigger spatial scale is called the \textit{reservoir scale} (Fig. \ref{fig_app_1}A). Regarding portions of matter, the reservoir scale is focused on instances of \ent{Portion of Rock} and its relations with other entities (Fig. \ref{fig_app_types}). A \textit{Reservoir Rock} is a \ent{Portion of Rock} that is porous and is a component of a \ent{Petroleum System}. In this application, we follow the conceptualization proposed by \cite{abel2015geomodeling}. In their work, they further analyze \textit{petroleum systems} and their components, and among them, a \ent{Reservoir Rock} is an essential part. \ent{Portion of Rock} is a type that specializes \ufo{Quantity} and is an instance of \ufo{Kind}. \ent{Reservoir Rock} is a type that specializes \ent{Portion of Rock} and, as an anti-rigid sortal type, that is relational dependent from a \ent{Petroleum System}, it is an instance of \ufo{Role}.

The second scale of observation, which we call the \textit{well core scale} (Fig. \ref{fig_app_1}B), is concerned with describing the entities related to a \ent{Well Core}, which is a cylindrical-shaped object with a radius in the order of $10^1$ cm, and a height of $10^1$-$10^2$ m. A \ent{Well Core} is an artifact extracted from the subsurface during the development of an \textit{oil field} to be used as a sample of the rocks in the field for some research or geologic analysis. The \ent{Well Core} type is a specialization of \ufo{Object} that instantiates \ufo{Kind}. A \ent{Well Core} is composed of some \ent{Sub-Portion of Rock}.

The third and last scale of observation, which we will call the \textit{thin section scale} (Fig. \ref{fig_app_1}C), is focused on a \ent{Thin Section}, which is an artifact with a thin slab of rock glued to a piece of glass. Thin sections are created with the intent of being analyzed through an optical microscope to allow the observation of the granules of the slab of rock. The \ent{Thin Section} type is a specialization of \ufo{Object} that instantiates \ufo{Kind}. A \ent{Thin Section} is composed of some \ent{Sub-Portion of Rock}. It is also composed of some piece of glass, that is not included in the model as it is not relevant right now.

\ent{Sedimentary Grain} is also targeted in the \textit{thin section scale}. A \ent{Sedimentary Grain} is a geological object created by a sedimentary process that eroded and transported a piece of another geological object or rock. \ent{Sedimentary Grain} is one of the possible types of object that can be granules of a \ent{Portion of Rock}. Then, \ent{Sedimentary Grain} type is a specialization of \ufo{Object} that instantiates \ufo{Kind}.

\begin{figure}
  \centering
  \includegraphics[width=\linewidth]{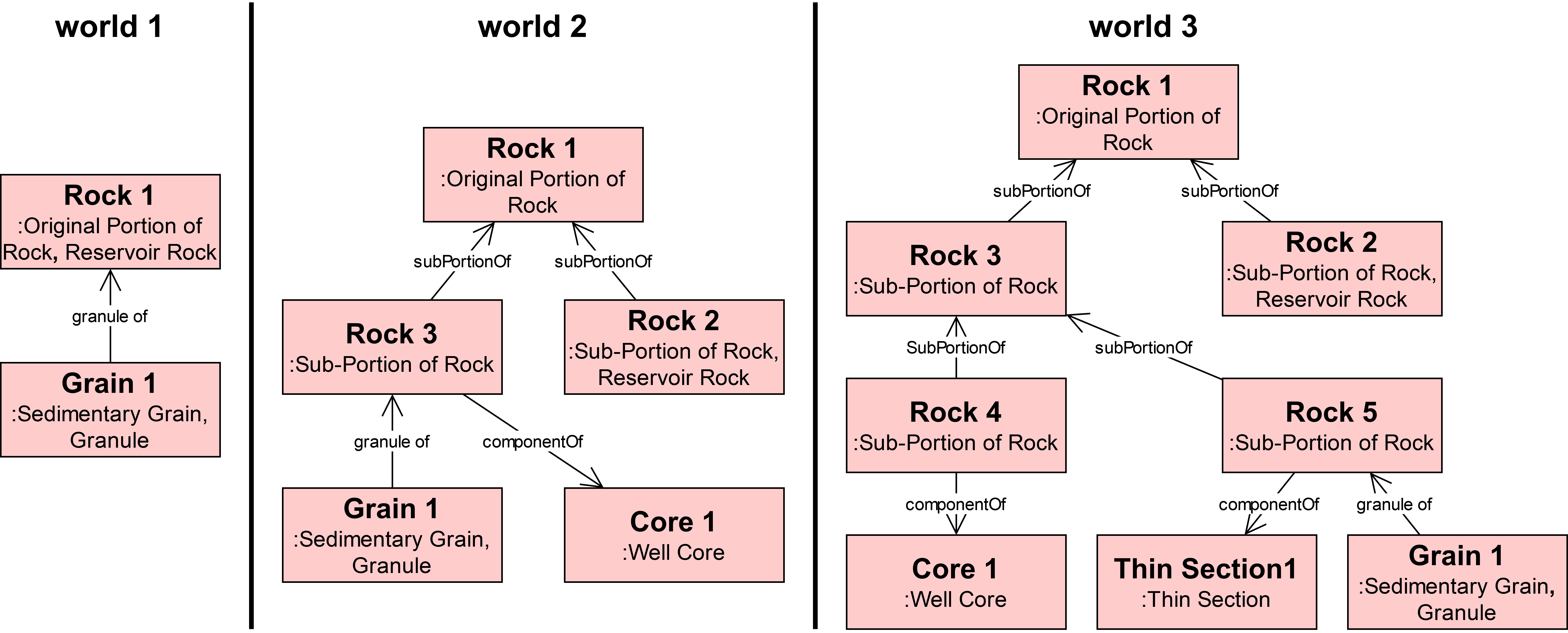}
  \caption{Object diagram showing three sequential worlds with the state of individuals at different time slices.}
  \label{fig_app_3}
\end{figure}

With the types and their relations defined so far, we can model the time slices between events that happened as \textit{worlds}. A \textit{world} is an extensional entity that is a maximal state of affairs, meaning that it contains all entities and relations in existence at the time. As we adopt the growing block view (Section \ref{section_basic_assumptions}), each everything in the past and in the present exist in each \textit{world}. Consequently, each event is preceded and succeeded by a situation that is part of some \textit{world}. With that, we model a concrete case using the above-mentioned scales of observation, using three worlds (Fig. \ref{fig_app_3}).

First, \textit{world 1} is the state of affairs before the events included in our model. In this world, \ent{rock 1} is a \ent{portion of rock} in the sub-surface, and it is a \ent{Reservoir Rock}, as it is porous and it is part of a petroleum system. Accordingly, \ent{rock 1} is an instance of \ent{Reservoir Rock}. Regarding if \ent{rock 1} is an \ent{Original Portion} or a \ent{Sub-Portion}, it can be an instance of any of these two types, but we will assume it to be an \ent{Original Portion}. Additionally, we will track \ent{grain 1}, that is an instance of \ent{Sedimentary Grain} and a granule of \ent{rock 1} in \textit{world 1}.

As the \ent{granuleOf} relation follows extensional mereology, \textit{rock 1} must have more than one granule. However, including all granules in the model is not viable, and most types unnecessary as they share the same provenance. But including some granules in the model already allow explicating and inferring knowledge. For this reason, We include only \ent{grain 1} as it suffices to demonstrate how to track the provenance of grains through time.

Second, \textit{world 2} is the state of affairs after a well-core extraction event. In this world, \textit{rock 1} was terminated (although it still exists with a historical nature), as it was split into two sub-portions, \ent{rock 2} and \ent{rock 3}. \ent{rock 2} remained in the subsurface and it assumes the \ent{Reservoir Rock} role previously taken by \textit{rock 1} in the petroleum system. \ent{rock 3} is the sub-portion of rock that is a component of \ent{core 1}, an instance of \ent{Well Core} which was also created during the well-core extraction event. 

\begin{figure}
  \centering
  \includegraphics[width=\linewidth]{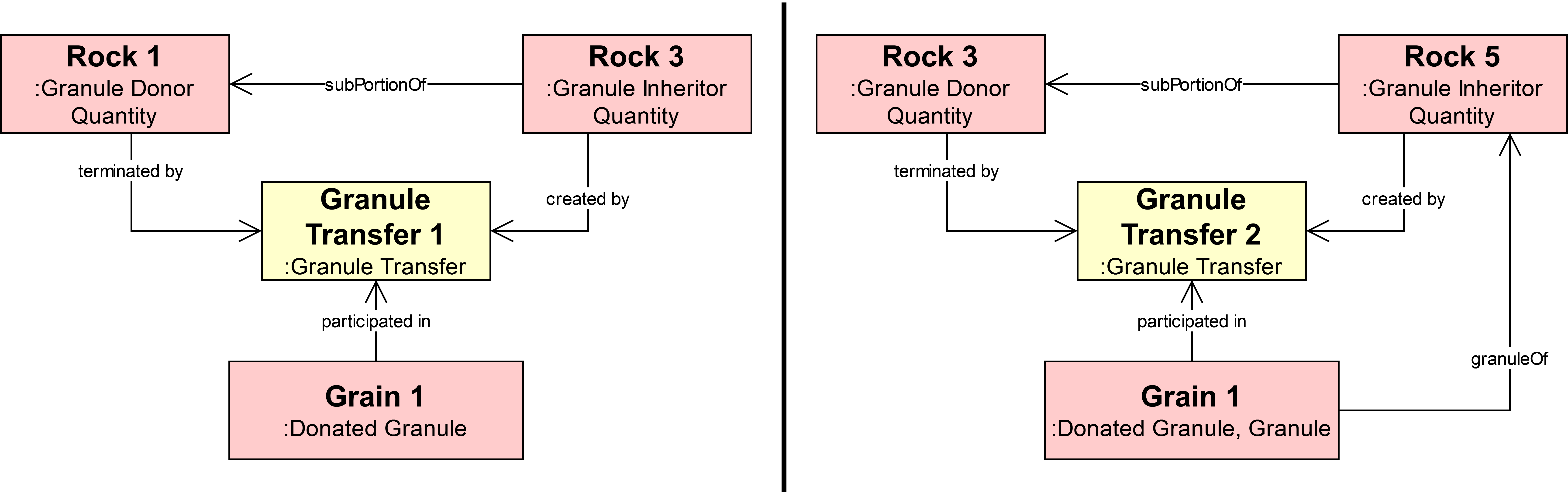}
  \caption{Object diagram with two intances of \ent{Granule Transfer} events in which \ent{grain 1} participated.}
  \label{fig_app_events}
\end{figure}

Regarding \ent{granule 1}, it is transferred from \ent{rock 1} to \ent{rock 3} during \ent{granule transfer 1} event (Fig. \ref{fig_app_events}A). This event follows the \ent{Granule Transfer} pattern (Fig. \ref{fig_subportion}), with \ent{rock 1} assuming the historical role of \ent{Granule Donor Quantity}, \ent{rock 3} assuming the role of \ent{Granule Inheritor Quantity}, and \ent{granule 1} assuming the role of \ent{Donated Granule}.

Third and last, \textit{world 3} is the state of affairs after the manufacturing of \ent{thin section 1}
by extracting a small portion of rock from \ent{core 1}. In this world, \ent{rock 3} was split into \ent{rock 4} and \ent{rock 5}. \ent{rock 4} replaces \ent{rock 3} as a component of \ent{core 1}. And \ent{rock 5} becomes a component of \ent{thin section 1}.

In \textit{world 3}, \ent{grain 1} was transferred from \ent{rock 3} to \ent{rock 5} during \ent{granule transfer 2} event (Fig. \ref{fig_app_events}B). Similarly to the previous event, \ent{rock 3} and \ent{rock 5} assume the roles of \ent{Granule Donor Quantity} and \ent{Granule Inheritor Quantity}, respectively, and \ent{granule 1} the role of \ent{Donated Granule}. 

With conceptual models like the one presented above, we can infer how the collection of granules of the sub-portion was inherited from other quantities by using the \ent{subPortionOf} relation. With that, it becomes explicit how matter is relocated from one portion of matter to another over time.

Another benefit of using the proposed ontology is that we can better capture information from geologists (the domain experts) by modeling relations through time. For example, a geologist can study \ent{grain 1} as a part of \ent{thin section 1} and identify events that changed it during the time it was a granule of \ent{rock 1}. So we can infer that the same events changed granules of \ent{rock 2}, the actual \ent{Reservoir Rock}, as they were all granules of \ent{rock 1} when the events happened.

This kind of reasoning is commonly implicit in the minds of the experts. Building information systems using an ontology-based conceptual model, can help attribute data more precisely to the entities. Furthermore, it allows the use of inference engines to extract more information and better find inconsistencies in data.

Furthermore, it also becomes clearer why some qualities of distinct sub-portions may have different values despite being historically related. If the qualities of the original portion of matter have some degree of heterogeneity, its sub-portions might inherit a distinct value for those qualities. For instance, a rock's permeability (a quality) depends on its granules' size. Then, using the previous example, if \ent{rock 1} is constituted by a collection of granules with distinct sizes, when it is split into \ent{rock 2} and \ent{rock 3}, one of these sub-portions might inherit a collection with larger granules then the other. Consequently, they will have permeability values different from each other and possibly different from \ent{rock 1} too.

\section{Related Work}\label{section_related_work}

Most top-level ontologies have a way of representing matter (mass nouns, material substance, or stuff) with distinct ontological commitments. GFO \citep{herre2010general} conceptualizes \textit{material substances} as instances of \textit{stuff}, which are continuants and necessarily dependent on \textit{Material Aggregates} (collections of \textit{Material Objects}). Every \textit{stuff} is part of the \textit{stuff} of some \textit{Material Aggregate}. The \textit{consistsOf} relation holds between a material object and its stuff.

BFO \citep{arp2015building} does not have a specific type in its basic taxonomy to categorize material substances. In fact, \cite{arp2015building} recommend not using \textit{mass nouns}, considering that their instances are not countable. Instead, they suggest attaching ``\textit{portion of}'' to the mass nouns to better transparency in the meaning of the concepts.

DOLCE \citep{gangemi2002sweetening} uses the notion of \textit{Amount of Matter} for instances of \textit{material substances}. \textit{Amounts of Matter} are mereologically invariant and do not have any unity criterion associated with them \citep{gangemi2002sweetening}.

Among some other relevant contributions on portions of matter, \cite{hahmann2014voids} discuss \textit{material constitution} throughout the granularity levels of materials and how it is related to voids. They formalize their contributions by aligning them with DOLCE. They use the notion of material-spatial interdependence \citep{hahmann2014interdependence} to define a set of relations between objects that share the same matter and have overlapping spatial regions. Hence, they define the \textit{submaterial} and \textit{maximal submaterial} relations that hold between two material entities that share their matter.

Regarding the \ent{submaterial} relation, it is a type of \textit{part of} relation restricted to material entities. In this sense, UFO provides more specific proper parthood relations in comparison to \ent{submaterial} relation. For instance, the \textit{componentOf} relation, if holding between material entities (\EG an engine and a car), would be a \ent{submaterial}. A \textit{subQuantityOf} relation (\EG between a portion of alcohol and a portion of wine) would also be a \ent{submaterial}. The \ent{granuleOf} relation proposed in our work would be another type of \ent{submaterial}.

The \ent{maximal submaterial} relation is a \ent{submaterial} relation in which the material entities share all their matter, but the occupied material region of one is a proper part of the other. This means that the larger region has voids (empty spaces) not included in the smaller region. For instance, a collection of $H_2O$ molecules is a \textit{maximal submaterial} of a portion of water, and the collection of hydrogen and oxygen atoms is also a \textit{maximal submaterial} of the portion of water, even if in a distinct level of granularity. Regarding this work, a \ent{Collection of Granules} would be the coarsest \ent{maximal submaterial} of a portion of matter. 

Finally, \cite{hahmann2014voids} define two kinds of \textit{constitution}:\textit{ intragranular constitution}, holding between entities in the same level of granularity (\EG a $H_2O$ molecule and the collection of molecules); and \textit{intergranular constitution}, between distinct levels of granularity (\EG a $H_2O$ molecule and the portion of water). Although these relations are useful to shed light on the distinct granularity levels, the conceptualization of \textit{material constitution} adopted by \cite{hahmann2014voids} is incompatible with this work as they disregard the need for complete co-location between constituted and constituent.

\cite{keet2014core} proposes a core ontology of macroscopic stuff, which they claim to be compatible with both DOLCE and BFO. This ontology contains a taxonomy of \textit{Stuff} focusing on the nature of the entities forming stuff on an immediate finer-grained level (granule, grain, or basis), on whether stuff is mixed, and on the nature of the mixture. \cite{keet2014core} uses the \textit{hasGranuleType} relation between \textit{stuff} and the type of object in the finer-grained level. This relation is equivalent in meaning to the one included in this work with the same name.

\cite{keet2016relating} further discusses parthood relations between amounts of matter, among which are \textit{stuff-part} between instances of distinct kinds and \textit{portion} between instances of the same kind. The \textit{stuff-part} relation is comparable with the UFO's \textit{subQuantityOf} relation. 

However, the \textit{portion} relation is conceptualized distinctly from how it is proposed in this work.
The \textit{portion} relation in \cite{keet2016relating} is mereological, holding between two instances of \textit{stuff} of the same kind. Differently from UFO and our ontology, they adopt the conceptualization of \textit{amount of matter} \cite[sensu][]{gangemi2002sweetening} for modeling \textit{stuff}. Hence, the portion relation holds, for instance, between a scattered stuff and each of its maximally-connected portions. However, UFO's quantities are maximally self-connected and consequently do not have parts of the same kind. Therefore, in our work, the \textit{sub-portion of} relation between quantities is not mereological and represents the notion that scattered portions of matter are usually connected to a previously existent portion of matter by historical relations.

Lastly, \cite{garcia2019rocks} uses \textit{material constitution} and mereological relations to propose a pattern to model rocks. The GeoCore Ontology of \cite{garcia2020geocore} is a core ontology for the geology domain built under BFO that integrates those ideas, including the \textit{constitutedBy} relation. Regarding matter, \cite{garcia2019rocks} adopt the conceptualization of \textit{amount of matter}  \citep[sensu][]{gangemi2002sweetening} to model \textit{earth materials} (a category of types of matter that include rock). They discuss how entities in the geology domain are related by \textit{material constitution}, including how the types of matter in the domain (e.g. rock and mineral) are constituted by collections of objects.

\section{Discussion}\label{section_discussion}

\subsection{Answering research questions}

We proposed using the \ent{granuleOf} parthood relation to answer the research question \textbf{RQ1} (\IE \textit{What is the meronymic relation between a portion of matter and its discrete parts?}), which concerns the parthood relation between an object that is part of a portion of matter. The \ent{granuleOf} holds between an \ufo{Object}, as the part, and a \ufo{Quantity}, as a whole. We also propose the type \ent{Granule} as the \ufo{RoleMixin} instantiated by objects whenever in this parthood relation.

Including this new meronymic relation, we can satisfactorily model that any \textit{stuff} is ultimately composed of objects. This includes certain materials with a highly homogeneous character (even when observed through a microscope), such as volcanic glass and water. This addition allows more precise and complete ontology-based knowledge models.

Secondly, we answer the research question \textbf{RQ2} (\IE \textit{What constitutes portions of matter?}), which concerns the constitution of portions of matter. We propose that \textit{quantities} are constituted by \textit{collections of granules}, which are collectives of adequate types of objects that, whenever in favorable circumstances, support the emergence of the characteristic properties of portions of matter and the consequent coming into existence of such portions. It should be noted that, although different types of quantities are associated with different favorable circumstances for their collection of granules, \textit{being physically connected} is one of such circumstances for all types of quantity.  

Thirdly, we answer \textbf{RQ3} (\IE \textit{How portions of matter are historically related through their discrete parts?}), which refers to how portions of matter are historically related through their discrete parts. This research question is motivated by the need to model events that affect materials, especially events that split, mix, or partially destroy them. Although it is possible to solve the issue simply using historical relations in an attributive manner, our ontology better explains these relations by using the \ent{granule transfer} and \ent{sub-portion formation} events as their truthmaker and making it explicit that matter might be transferred through granules. As a consequence, the historical relations \ent{inheritsGranulesFrom}, \ent{donatedGranulesTo}, and \ent{subPortionOf} (Sections \ref{section_granule_inheritance} and \ref{section_subportion}) are better grounded by these events.

\subsection{Are all quantities composed of granules?}

As UFO's \textit{quantities} are defined, we understand that they only encompass \textit{stuff}. However, further foundational research into material entities is still needed to understand whether there are portions of matter not constituted by objects. For example, it is still not fully understood the nature of the matter that constitutes sub-atomic particles. Also, it is not clear whether there are other types of relations beyond physical connection that could be considered to cause the emergence of quantities. For instance, we could conceptualize cosmic dust as a type of portion of matter, although its granules are connected through a gravitational relation instead of a physical connection.

Most domains, however, deal with matter entities on an intermediate scale compared to small and large-scale physics. For these domains, it seems reasonable to consider \textit{quantities} as portions of \textit{stuff}, and consequently, constituted of collections of granules.

\subsection{Portions of matter as collections}

Another point of discussion refers to the possible conceptualization of \textit{portions of matter} as collectives. In other words, since portions of matter are formed by physically connected discrete parts, why not simply equate \textit{portions of matter} to such a collection of discrete parts instead of requiring an additional, constituted entity (the portion)? For instance, according to this view, a portion of water is nothing more than a collection of $H_2O$ molecules. 

Although such an ontological decision has nothing wrong \textit{per se}, we defend that a collection of granules and a quantity are distinct entities with different causal powers. The structure and interactions between the granules are the basis of the emergent properties of a quantity and essential for its identity, whereas collectives are unified solely by some characterizing relation \cite{guizzardi2011collectives}. In this sense, quantities resemble functional complexes as there is some sort of ``functionality'' in the relation between the parts while resembling collectives by having a part-whole relation that is also uniform.

To exemplify the importance of interactions between parts to a quantity we can resort to the notion of ``crowd'', which is usually conceptualized as a collective. In regular settings, a crowd (as a collective) is unified by the relation of its members being in the same place. But in a scenario in which the crowd is dense enough, and some disturbance occurs, the crowd will begin to have a fluid-like behavior as people start to bump into each other and move around obstacles. It would be reasonable to model that crowd as a quantity in such a scenario. This is very similar to regular stuff, but instead of people bumping into each other, it might be sedimentary grains or molecules with distinct types of interaction between them.

Furthermore, quantities have different moments from their constituent collections. For instance, a collection of carbon atoms can constitute a portion of diamond or a portion of graphite. While a portion of graphite is very soft and has high electric conductivity, the portion of diamond is one of the most hard natural materials, with low electric conductivity. In such an example, hardness and electrical conductivity are not characteristics of collections of atoms, but of the type of matter emerging after the favorable conditions are met.

\subsection{Granularity levels and negative parts}
In the proposed \textit{ontology of portions of matter}, we have only considered the relations of granules and quantities on linking only two levels of granularity, \IE the level of the quantity and the level of the granule. However, in nature, matter might be organized in several levels of granularity, and expanding the ontology to account for those levels of granularity and their change is desirable. For instance, a rock might have granules that are constituted of a portion of a second rock type, and this portion have granules constituted of a third rock type.

Secondly, we did not consider the changes that granules might go through while still being parts of a quantity and the implications of those changes associated with granularity levels. For example, atoms that are granules of certain material (\EG a mineral) might decay naturally, causing changes in that mineral, but also on a rock that has a granule constituted of that mineral. 

Lastly, our proposed \textit{ontology of portions of matter} also does not include holes (negative parts, \EG pores in a rock), to represent the non-material parts of quantities. Holes can be useful in an ontology of portion of matters for two reasons, first to aid in further distinguishing between quantities and the collectives constituting them. For instance, pores can be considered to be part of a rock; however, they are not part of the collection of granules. Second, holes also have an important relation to specific modes of portions of matter. For instance, the porosity of a rock is a quality grounded in the holes between the rock's granules. Therefore, future work should include a representation of holes in the \textit{ontology of portions of matter}. 

\section{Conclusion}\label{section_conclusion}

This paper presented a complementary ontological analysis of portions of matter under UFO top-level ontology, providing a more in-depth discussion about portions of matter (\IE ``quantities'' in UFO). In this work, three main contributions are found. First, the proposal of the \textit{granuleOf} meronymic relation, holding between a functional complex and a quantity. Second, we discuss the constitution of quantities proposing \textit{collection of granules} as their constituents. Third, we propose the \textit{inheritsGranulesFrom} and \textit{subPortionOf} historical relations between portions of matter that are grounded on granule transfer events.

We also apply the proposed ontology to a case study in the Oil \& Gas industry (section \ref{section_case_study}), illustrating the usage of the ontology. Although motivated by research questions derived from issues found in information systems in the geology and oil \& gas domain, the development of this ontological analysis is not limited in application to these domains, but can be useful for areas such as medicine, material engineering, the food industry and others that deal with matter entities.  

In future work, we intend to extend the proposed ontology in three directions. The first is the addition of granularity levels between objects. Another is the definition of a taxonomy of events that create and terminate quantities. Lastly, we will investigate the negative parts (holes) and their relation to quantities.

In closing, our analysis of portions of matter has not only deepened our understanding of their properties and relations but has also highlighted the significance of continued research in ontologies for portions of matter. The knowledge gained from this study serves as a stepping stone for future research, potentially unlocking further discoveries and advancements.

\begin{acks}
Lucas Valadares Vieira thanks Petrobras for supporting this work. Mara Abel acknowledges CAPES-Brazil Finance Code 001, the Brazilian Agency CNPq, and the Georeservoir project supported by Petrobras. Fabricio Rodrigues thanks the Petwin Project, supported by FINEP, Libra Consortium (Petrobras, Shell Brasil, Total Energies, CNOOC, CNPC).  
\end{acks}

\nocite{*}
\bibliographystyle{ios2-nameyear}  
\bibliography{bibliography}        

\begin{thebibliography}{48}
\ifx \bisbn   \undefined \def \bisbn  #1{ISBN #1}\fi
\ifx \binits  \undefined \def \binits#1{#1} \fi
\ifx \bauthor  \undefined \def \bauthor#1{#1} \fi
\ifx \bjtitle  \undefined \def \bjtitle#1{\textit{#1}}\fi
\ifx \batitle  \undefined \def \batitle#1{#1} \fi
\ifx \bctitle  \undefined \def \bctitle#1{#1} \fi
\ifx \bvolume  \undefined \def \bvolume#1{#1}\fi
\ifx \byear  \undefined \def \byear#1{#1} \fi
\ifx \bissue  \undefined \def \bissue#1{#1} \fi
\ifx \bfpage  \undefined \def \bfpage#1{#1} \fi
\ifx \blpage  \undefined \def \blpage #1{#1} \fi
\ifx \url  \undefined \def \url#1{#1} \fi
\ifx \doiurl  \undefined \def \doiurl#1{#1} \fi
\ifx \betal  \undefined \def \betal{et al.} \fi
\ifx \binstitute  \undefined \def \binstitute#1{#1} \fi
\ifx \beditor  \undefined \def \beditor#1{#1} \fi
\ifx \bpublisher  \undefined \def \bpublisher#1{#1} \fi
\ifx \bbtitle  \undefined \def \bbtitle#1{\textit{#1}} \fi
\ifx \bedition  \undefined \def \bedition#1{#1} \fi
\ifx \bseriesno  \undefined \def \bseriesno#1{\textit{#1}} \fi
\ifx \blocation  \undefined \def \blocation#1{#1} \fi
\ifx \bsertitle  \undefined \def \bsertitle#1{\textit{#1}} \fi
\ifx \bsnm \undefined \def \bsnm#1{#1} \fi
\ifx \bsuffix \undefined \def \bsuffix#1{#1} \fi
\ifx \bparticle \undefined \def \bparticle#1{#1} \fi
\ifx \barticle \undefined \def \barticle#1{#1} \fi
\ifx \botherref \undefined \def \botherref #1{#1} \fi
\ifx \url \undefined \def \url#1{#1} \fi
\ifx \bchapter \undefined \def \bchapter#1{#1} \fi
\ifx \bbook \undefined \def \bbook#1{#1} \fi
\ifx \bcomment \undefined \def \bcomment#1{#1} \fi
\ifx \oauthor \undefined \def \oauthor#1{#1} \fi
\ifx \citeauthoryear \undefined \def \citeauthoryear#1{#1} \fi
\ifx \bconflocation \undefined \def \bconflocation#1{#1} \fi
\ifx \texttildelow  \undefined \def \texttildelow{\symbol{126}} \fi
\def \endbibitem {}

\bibitem[\protect\citeauthoryear{Abel et~al.}{2015}]{abel2015geomodeling}
\begin{barticle}
\bauthor{\bsnm{Abel}, \binits{M.}},
\bauthor{\bsnm{Perrin}, \binits{M.}} \&
\bauthor{\bsnm{Carbonera}, \binits{J.L.}}
(\byear{2015}).
\batitle{Ontological analysis for information integration in geomodeling}.
\bjtitle{Earth Science Informatics},
\bvolume{8},
\bfpage{21}--\blpage{36}.
\end{barticle}
\endbibitem

\bibitem[\protect\citeauthoryear{Abel et~al.}{2004}]{abel2004petrographer}
\begin{barticle}
\bauthor{\bsnm{Abel}, \binits{M.}},
\bauthor{\bsnm{Silva}, \binits{L.A.}},
\bauthor{\bsnm{De~Ros}, \binits{L.F.}},
\bauthor{\bsnm{Mastella}, \binits{L.S.}},
\bauthor{\bsnm{Campbell}, \binits{J.A.}} \&
\bauthor{\bsnm{Novello}, \binits{T.}}
(\byear{2004}).
\batitle{PetroGrapher: managing petrographic data and knowledge using an
  intelligent database application}.
\bjtitle{Expert Systems with Applications},
\bvolume{26}(\bissue{1}),
\bfpage{9}--\blpage{18}.
\end{barticle}
\endbibitem

\bibitem[\protect\citeauthoryear{Abel et~al.}{2015}]{abel2015maturedomain}
\begin{bchapter}
\bauthor{\bsnm{Abel}, \binits{M.}},
\bauthor{\bsnm{Carbonera}, \binits{J.L.}},
\bauthor{\bsnm{Fiorini}, \binits{S.R.}},
\bauthor{\bsnm{Garcia}, \binits{L.F.}} \&
\bauthor{\bsnm{De~Ros}, \binits{L.F.}}
(\byear{2015}).
\bctitle{The Multiple Applications of a Mature Domain Ontology.}
In \bbtitle{ONTOBRAS}.
\end{bchapter}
\endbibitem

\bibitem[\protect\citeauthoryear{Almeida et~al.}{2019a}]{almeida2019events}
\begin{bchapter}
\bauthor{\bsnm{Almeida}, \binits{J.P.A.}},
\bauthor{\bsnm{Falbo}, \binits{R.A.}} \&
\bauthor{\bsnm{Guizzardi}, \binits{G.}}
(\byear{2019}a).
\bctitle{Events as entities in ontology-driven conceptual modeling}.
In \bbtitle{Conceptual Modeling: 38th International Conference, ER 2019,
  Salvador, Brazil, November 4--7, 2019, Proceedings 38}
(pp. \bfpage{469}--\blpage{483}).
\binstitute{Springer}.
\end{bchapter}
\endbibitem

\bibitem[\protect\citeauthoryear{Almeida
  et~al.}{2017}]{almeida2017comprehensive}
\begin{bchapter}
\bauthor{\bsnm{Almeida}, \binits{J.P.A.}},
\bauthor{\bsnm{Fonseca}, \binits{C.M.}} \&
\bauthor{\bsnm{Carvalho}, \binits{V.A.}}
(\byear{2017}).
\bctitle{A comprehensive formal theory for multi-level conceptual modeling}.
In \bbtitle{Conceptual Modeling: 36th International Conference, ER 2017,
  Valencia, Spain, November 6--9, 2017, Proceedings 36}
(pp. \bfpage{280}--\blpage{294}).
\binstitute{Springer}.
\end{bchapter}
\endbibitem

\bibitem[\protect\citeauthoryear{Almeida et~al.}{2019b}]{almeida2019gufo}
\begin{botherref}
\oauthor{\bsnm{Almeida}, \binits{J.}},
\oauthor{\bsnm{Guizzardi}, \binits{G.}},
\oauthor{\bsnm{Falbo}, \binits{R.}} \&
\oauthor{\bsnm{Sales}, \binits{T.P.}}
(2019b).
gUFO: a lightweight implementation of the Unified Foundational Ontology (UFO).
\textit{URL http://purl. org/nemo/doc/gufo}.
\end{botherref}
\endbibitem

\bibitem[\protect\citeauthoryear{Arp et~al.}{2015}]{arp2015building}
\begin{bbook}
\bauthor{\bsnm{Arp}, \binits{R.}},
\bauthor{\bsnm{Smith}, \binits{B.}} \&
\bauthor{\bsnm{Spear}, \binits{A.D.}}
(\byear{2015}).
\bbtitle{Building ontologies with basic formal ontology}.
\bpublisher{Mit Press}.
\end{bbook}
\endbibitem

\bibitem[\protect\citeauthoryear{Baker}{2000}]{baker2000persons}
\begin{bbook}
\bauthor{\bsnm{Baker}, \binits{L.R.}}
(\byear{2000}).
\bbtitle{Persons and bodies: A constitution view}.
\bpublisher{Cambridge University Press}.
\end{bbook}
\endbibitem

\bibitem[\protect\citeauthoryear{Beckett et~al.}{2014}]{beckett2014rdf}
\begin{botherref}
\oauthor{\bsnm{Beckett}, \binits{D.}},
\oauthor{\bsnm{Berners-Lee}, \binits{T.}},
\oauthor{\bsnm{Prud’hommeaux}, \binits{E.}} \&
\oauthor{\bsnm{Carothers}, \binits{G.}}
(2014).
RDF 1.1 Turtle.
\textit{World Wide Web Consortium},
18--31.
\end{botherref}
\endbibitem

\bibitem[\protect\citeauthoryear{Britannica}{n.d.}]{BritannicaMatter}
\begin{botherref}
\oauthor{\bsnm{Britannica}}
(n.d.).
Matter.
Encyclopædia Britannica, inc.
\url{https://www.britannica.com/science/matter}.
\end{botherref}
\endbibitem

\bibitem[\protect\citeauthoryear{Burek et~al.}{2020}]{burek2020gfo2}
\begin{bchapter}
\bauthor{\bsnm{Burek}, \binits{P.}},
\bauthor{\bsnm{Loebe}, \binits{F.}} \&
\bauthor{\bsnm{Herre}, \binits{H.}}
(\byear{2020}).
\bctitle{Towards GFO 2.0: Architecture, modules and applications}.
In \bbtitle{Formal Ontology in Information Systems}
(pp. \bfpage{32}--\blpage{45}).
\binstitute{IOS Press}.
\end{bchapter}
\endbibitem

\bibitem[\protect\citeauthoryear{Carvalho et~al.}{2017}]{carvalho2017multi}
\begin{barticle}
\bauthor{\bsnm{Carvalho}, \binits{V.A.}},
\bauthor{\bsnm{Almeida}, \binits{J.P.A.}},
\bauthor{\bsnm{Fonseca}, \binits{C.M.}} \&
\bauthor{\bsnm{Guizzardi}, \binits{G.}}
(\byear{2017}).
\batitle{Multi-level ontology-based conceptual modeling}.
\bjtitle{Data \& Knowledge Engineering},
\bvolume{109},
\bfpage{3}--\blpage{24}.
\end{barticle}
\endbibitem

\bibitem[\protect\citeauthoryear{Cicconeto et~al.}{2020}]{cicconeto2020spatial}
\begin{bchapter}
\bauthor{\bsnm{Cicconeto}, \binits{F.}},
\bauthor{\bsnm{Vieira}, \binits{L.V.}},
\bauthor{\bsnm{Abel}, \binits{M.}},
\bauthor{\bsnm{dos Santos~Alvarenga}, \binits{R.}} \&
\bauthor{\bsnm{Carbonera}, \binits{J.L.}}
(\byear{2020}).
\bctitle{A Spatial Relation Ontology for Deep-Water Depositional System
  Description in Geology.}
In \bbtitle{ONTOBRAS}
(pp. \bfpage{35}--\blpage{47}).
\end{bchapter}
\endbibitem

\bibitem[\protect\citeauthoryear{Dodds and Davis}{2011}]{dodds2011linked}
\begin{bbook}
\bauthor{\bsnm{Dodds}, \binits{L.}} \&
\bauthor{\bsnm{Davis}, \binits{I.}}
(\byear{2011}).
\bbtitle{Linked data patterns}.
\end{bbook}
\endbibitem

\bibitem[\protect\citeauthoryear{Donnelly}{2016}]{maureen2016threedimensionalism}
\begin{bchapter}
\bauthor{\bsnm{Donnelly}, \binits{M.}}
(\byear{2016}).
\bctitle{{Three-Dimensionalism}}.
In \bbtitle{{The Oxford Handbook of Topics in Philosophy}}.
\bpublisher{Oxford University Press}.
doi:\doiurl{10.1093/oxfordhb/9780199935314.013.39}.
\end{bchapter}
\endbibitem

\bibitem[\protect\citeauthoryear{Emery et~al.}{2020}]{Emery2020time}
\begin{bchapter}
\bauthor{\bsnm{Emery}, \binits{N.}},
\bauthor{\bsnm{Markosian}, \binits{N.}} \&
\bauthor{\bsnm{Sullivan}, \binits{M.}}
(\byear{2020}).
\bctitle{{Time}}.
In \beditor{\binits{E.N.} \bsnm{Zalta}} (Ed.),
\bbtitle{The {Stanford} Encyclopedia of Philosophy}
(\bedition{{W}inter 2020} ed.).
\bpublisher{Metaphysics Research Lab, Stanford University}.
\end{bchapter}
\endbibitem

\bibitem[\protect\citeauthoryear{Fernandes}{2020}]{fernandes2020clustering}
\begin{botherref}
\oauthor{\bsnm{Fernandes}, \binits{L.P.}}
(2020).
A clustering-based approach to identify petrofacies from petrographic data.
\end{botherref}
\endbibitem

\bibitem[\protect\citeauthoryear{Fine}{1999}]{fine1999things}
\begin{barticle}
\bauthor{\bsnm{Fine}, \binits{K.}}
(\byear{1999}).
\batitle{Things and their parts}.
\bjtitle{Midwest Studies in Philosophy},
\bvolume{23},
\bfpage{61}--\blpage{74}.
\end{barticle}
\endbibitem

\bibitem[\protect\citeauthoryear{Fonseca et~al.}{2019}]{fonseca2019relations}
\begin{bchapter}
\bauthor{\bsnm{Fonseca}, \binits{C.M.}},
\bauthor{\bsnm{Porello}, \binits{D.}},
\bauthor{\bsnm{Guizzardi}, \binits{G.}},
\bauthor{\bsnm{Almeida}, \binits{J.P.A.}} \&
\bauthor{\bsnm{Guarino}, \binits{N.}}
(\byear{2019}).
\bctitle{Relations in ontology-driven conceptual modeling}.
In \bbtitle{Conceptual Modeling: 38th International Conference, ER 2019,
  Salvador, Brazil, November 4--7, 2019, Proceedings 38}
(pp. \bfpage{28}--\blpage{42}).
\binstitute{Springer}.
\end{bchapter}
\endbibitem

\bibitem[\protect\citeauthoryear{Fonseca
  et~al.}{2022}]{fonseca2022incorporating}
\begin{bchapter}
\bauthor{\bsnm{Fonseca}, \binits{C.M.}},
\bauthor{\bsnm{Guizzardi}, \binits{G.}},
\bauthor{\bsnm{Almeida}, \binits{J.P.A.}},
\bauthor{\bsnm{Sales}, \binits{T.P.}} \&
\bauthor{\bsnm{Porello}, \binits{D.}}
(\byear{2022}).
\bctitle{Incorporating Types of Types in Ontology-Driven Conceptual Modeling}.
In \bbtitle{International Conference on Conceptual Modeling}
(pp. \bfpage{18}--\blpage{34}).
\binstitute{Springer}.
\end{bchapter}
\endbibitem

\bibitem[\protect\citeauthoryear{Galton}{2018}]{galton2018yet}
\begin{bchapter}
\bauthor{\bsnm{Galton}, \binits{A.}}
(\byear{2018}).
\bctitle{Yet Another Taxonomy of Part-Whole Relations.}
In \bbtitle{JOWO}.
\end{bchapter}
\endbibitem

\bibitem[\protect\citeauthoryear{Gangemi et~al.}{2002}]{gangemi2002sweetening}
\begin{bchapter}
\bauthor{\bsnm{Gangemi}, \binits{A.}},
\bauthor{\bsnm{Guarino}, \binits{N.}},
\bauthor{\bsnm{Masolo}, \binits{C.}},
\bauthor{\bsnm{Oltramari}, \binits{A.}} \&
\bauthor{\bsnm{Schneider}, \binits{L.}}
(\byear{2002}).
\bctitle{Sweetening ontologies with DOLCE}.
In \bbtitle{International conference on knowledge engineering and knowledge
  management}
(pp. \bfpage{166}--\blpage{181}).
\binstitute{Springer}.
\end{bchapter}
\endbibitem

\bibitem[\protect\citeauthoryear{Garcia et~al.}{2019}]{garcia2019rocks}
\begin{bchapter}
\bauthor{\bsnm{Garcia}, \binits{L.F.}},
\bauthor{\bsnm{Carbonera}, \binits{J.L.}},
\bauthor{\bsnm{Rodrigues}, \binits{F.H.}},
\bauthor{\bsnm{Antunes}, \binits{C.R.}} \&
\bauthor{\bsnm{Abel}, \binits{M.}}
(\byear{2019}).
\bctitle{What rocks are made of: towards an ontological pattern for material
  constitution in the geological domain}.
In \bbtitle{Conceptual Modeling: 38th International Conference, ER 2019,
  Salvador, Brazil, November 4--7, 2019, Proceedings 38}
(pp. \bfpage{275}--\blpage{286}).
\binstitute{Springer}.
\end{bchapter}
\endbibitem

\bibitem[\protect\citeauthoryear{Garcia et~al.}{2020}]{garcia2020geocore}
\begin{barticle}
\bauthor{\bsnm{Garcia}, \binits{L.F.}},
\bauthor{\bsnm{Abel}, \binits{M.}},
\bauthor{\bsnm{Perrin}, \binits{M.}} \&
\bauthor{\bsnm{dos Santos~Alvarenga}, \binits{R.}}
(\byear{2020}).
\batitle{The GeoCore ontology: a core ontology for general use in geology}.
\bjtitle{Computers \& Geosciences},
\bvolume{135},
\bfpage{104387}.
\end{barticle}
\endbibitem

\bibitem[\protect\citeauthoryear{Guizzardi}{2005}]{guizzardi2005}
\begin{bbook}
\bauthor{\bsnm{Guizzardi}, \binits{G.}}
(\byear{2005}).
\bbtitle{Ontological foundations for structural conceptual models}.
\end{bbook}
\endbibitem

\bibitem[\protect\citeauthoryear{Guizzardi}{2010}]{guizzardi2010}
\begin{bchapter}
\bauthor{\bsnm{Guizzardi}, \binits{G.}}
(\byear{2010}).
\bctitle{On the Representation of Quantities and their Parts in Conceptual
  Modeling.}
In \bbtitle{FOIS}
(pp. \bfpage{103}--\blpage{116}).
\end{bchapter}
\endbibitem

\bibitem[\protect\citeauthoryear{Guizzardi}{2011}]{guizzardi2011collectives}
\begin{bchapter}
\bauthor{\bsnm{Guizzardi}, \binits{G.}}
(\byear{2011}).
\bctitle{Ontological foundations for conceptual part-whole relations: The case
  of collectives and their parts}.
In \bbtitle{Advanced Information Systems Engineering: 23rd International
  Conference, CAiSE 2011, London, UK, June 20-24, 2011. Proceedings 23}
(pp. \bfpage{138}--\blpage{153}).
\binstitute{Springer}.
\end{bchapter}
\endbibitem

\bibitem[\protect\citeauthoryear{Guizzardi et~al.}{2013}]{guizzardi2013events}
\begin{bchapter}
\bauthor{\bsnm{Guizzardi}, \binits{G.}},
\bauthor{\bsnm{Wagner}, \binits{G.}},
\bauthor{\bsnm{de Almeida~Falbo}, \binits{R.}},
\bauthor{\bsnm{Guizzardi}, \binits{R.S.}} \&
\bauthor{\bsnm{Almeida}, \binits{J.P.A.}}
(\byear{2013}).
\bctitle{Towards ontological foundations for the conceptual modeling of
  events}.
In \bbtitle{Conceptual Modeling: 32th International Conference, ER 2013,
  Hong-Kong, China, November 11-13, 2013. Proceedings 32}
(pp. \bfpage{327}--\blpage{341}).
\binstitute{Springer}.
\end{bchapter}
\endbibitem

\bibitem[\protect\citeauthoryear{Guizzardi et~al.}{2022}]{guizzardi2022ufo}
\begin{barticle}
\bauthor{\bsnm{Guizzardi}, \binits{G.}},
\bauthor{\bsnm{Botti~Benevides}, \binits{A.}},
\bauthor{\bsnm{Fonseca}, \binits{C.M.}},
\bauthor{\bsnm{Porello}, \binits{D.}},
\bauthor{\bsnm{Almeida}, \binits{J.P.A.}} \&
\bauthor{\bsnm{Prince~Sales}, \binits{T.}}
(\byear{2022}).
\batitle{UFO: Unified foundational ontology}.
\bjtitle{Applied ontology},
\bvolume{17}(\bissue{1}),
\bfpage{167}--\blpage{210}.
\end{barticle}
\endbibitem

\bibitem[\protect\citeauthoryear{Hahmann and Brodaric}{2014}]{hahmann2014voids}
\begin{bchapter}
\bauthor{\bsnm{Hahmann}, \binits{T.}} \&
\bauthor{\bsnm{Brodaric}, \binits{B.}}
(\byear{2014}).
\bctitle{Voids and material constitution across physical granularities.}
In \bbtitle{FOIS}
(pp. \bfpage{51}--\blpage{64}).
\end{bchapter}
\endbibitem

\bibitem[\protect\citeauthoryear{Hahmann
  et~al.}{2014}]{hahmann2014interdependence}
\begin{bchapter}
\bauthor{\bsnm{Hahmann}, \binits{T.}},
\bauthor{\bsnm{Brodaric}, \binits{B.}} \&
\bauthor{\bsnm{Gr{\"u}ninger}, \binits{M.}}
(\byear{2014}).
\bctitle{Interdependence among material objects and voids}.
In \bbtitle{Formal Ontology in Information Systems}
(pp. \bfpage{37}--\blpage{50}).
\binstitute{IOS Press}.
\end{bchapter}
\endbibitem

\bibitem[\protect\citeauthoryear{Herre}{2010}]{herre2010general}
\begin{bchapter}
\bauthor{\bsnm{Herre}, \binits{H.}}
(\byear{2010}).
\bctitle{General Formal Ontology (GFO): A foundational ontology for conceptual
  modelling}.
In \bbtitle{Theory and applications of ontology: computer applications}
(pp. \bfpage{297}--\blpage{345}).
\bpublisher{Springer}.
\end{bchapter}
\endbibitem

\bibitem[\protect\citeauthoryear{Hitzler et~al.}{2009}]{hitzler2009owl}
\begin{barticle}
\bauthor{\bsnm{Hitzler}, \binits{P.}},
\bauthor{\bsnm{Kr{\"o}tzsch}, \binits{M.}},
\bauthor{\bsnm{Parsia}, \binits{B.}},
\bauthor{\bsnm{Patel-Schneider}, \binits{P.F.}},
\bauthor{\bsnm{Rudolph}, \binits{S.}}, \betal\
(\byear{2009}).
\batitle{OWL 2 web ontology language primer}.
\bjtitle{W3C recommendation},
\bvolume{27}(\bissue{1}),
\bfpage{123}.
\end{barticle}
\endbibitem

\bibitem[\protect\citeauthoryear{Horrocks et~al.}{2004}]{horrocks2004swrl}
\begin{barticle}
\bauthor{\bsnm{Horrocks}, \binits{I.}},
\bauthor{\bsnm{Patel-Schneider}, \binits{P.F.}},
\bauthor{\bsnm{Boley}, \binits{H.}},
\bauthor{\bsnm{Tabet}, \binits{S.}},
\bauthor{\bsnm{Grosof}, \binits{B.}},
\bauthor{\bsnm{Dean}, \binits{M.}}, \betal\
(\byear{2004}).
\batitle{SWRL: A semantic web rule language combining OWL and RuleML}.
\bjtitle{W3C Member submission},
\bvolume{21}(\bissue{79}),
\bfpage{1}--\blpage{31}.
\end{barticle}
\endbibitem

\bibitem[\protect\citeauthoryear{Jackson}{2012}]{jackson2012alloy}
\begin{bbook}
\bauthor{\bsnm{Jackson}, \binits{D.}}
(\byear{2012}).
\bbtitle{Software Abstractions: logic, language, and analysis}.
\bpublisher{MIT press}.
\end{bbook}
\endbibitem

\bibitem[\protect\citeauthoryear{Keet}{2014}]{keet2014core}
\begin{bchapter}
\bauthor{\bsnm{Keet}, \binits{C.M.}}
(\byear{2014}).
\bctitle{A core ontology of macroscopic stuff}.
In \bbtitle{International Conference on Knowledge Engineering and Knowledge
  Management}
(pp. \bfpage{209}--\blpage{224}).
\binstitute{Springer}.
\end{bchapter}
\endbibitem

\bibitem[\protect\citeauthoryear{Keet}{2016}]{keet2016relating}
\begin{bchapter}
\bauthor{\bsnm{Keet}, \binits{C.M.}}
(\byear{2016}).
\bctitle{Relating some stuff to other stuff}.
In \bbtitle{European Knowledge Acquisition Workshop}
(pp. \bfpage{368}--\blpage{383}).
\binstitute{Springer}.
\end{bchapter}
\endbibitem

\bibitem[\protect\citeauthoryear{Keet}{2017}]{keet2017note}
\begin{botherref}
\oauthor{\bsnm{Keet}, \binits{C.M.}}
(2017).
A note on the compatibility of part-whole relations with foundational
  ontologies.
\end{botherref}
\endbibitem

\bibitem[\protect\citeauthoryear{Loebe et~al.}{2022}]{loebe2022gfo}
\begin{barticle}
\bauthor{\bsnm{Loebe}, \binits{F.}},
\bauthor{\bsnm{Burek}, \binits{P.}} \&
\bauthor{\bsnm{Herre}, \binits{H.}}
(\byear{2022}).
\batitle{GFO: The General Formal Ontology}.
\bjtitle{Applied Ontology},
\bvolume{17}(\bissue{1}),
\bfpage{71}--\blpage{106}.
\end{barticle}
\endbibitem

\bibitem[\protect\citeauthoryear{Mastella et~al.}{2007}]{mastella2007event}
\begin{botherref}
\oauthor{\bsnm{Mastella}, \binits{L.S.}},
\oauthor{\bsnm{Abel}, \binits{M.}},
\oauthor{\bsnm{De~Ros}, \binits{L.F.}},
\oauthor{\bsnm{Perrin}, \binits{M.}} \&
\oauthor{\bsnm{Rainaud}, \binits{J.-F.}}
(2007).
Event ordering reasoning ontology applied to petrology and geological
  modelling.
\textit{Theoretical advances and applications of fuzzy logic and soft
  computing},
465--475.
\end{botherref}
\endbibitem

\bibitem[\protect\citeauthoryear{Mendon{\c{c}}a and
  Almeida}{2013}]{mendonca2013hemocomponents}
\begin{bchapter}
\bauthor{\bsnm{Mendon{\c{c}}a}, \binits{F.}} \&
\bauthor{\bsnm{Almeida}, \binits{M.B.}}
(\byear{2013}).
\bctitle{Hemocomponents and Hemoderivatives Ontology (HEMONTO): an Ontology
  About Blood Components.}
In \bbtitle{ONTOBRAS}
(pp. \bfpage{11}--\blpage{22}).
\end{bchapter}
\endbibitem

\bibitem[\protect\citeauthoryear{Mungall et~al.}{2012}]{mungall2012uberon}
\begin{barticle}
\bauthor{\bsnm{Mungall}, \binits{C.J.}},
\bauthor{\bsnm{Torniai}, \binits{C.}},
\bauthor{\bsnm{Gkoutos}, \binits{G.V.}},
\bauthor{\bsnm{Lewis}, \binits{S.E.}} \&
\bauthor{\bsnm{Haendel}, \binits{M.A.}}
(\byear{2012}).
\batitle{Uberon, an integrative multi-species anatomy ontology}.
\bjtitle{Genome Biology},
\bvolume{13},
\bfpage{R5}.
\url{https://api.semanticscholar.org/CorpusID:15453742}.
\end{barticle}
\endbibitem

\bibitem[\protect\citeauthoryear{Pan and Hobbs}{2006}]{pan2006time}
\begin{botherref}
\oauthor{\bsnm{Pan}, \binits{F.}} \&
\oauthor{\bsnm{Hobbs}, \binits{J.R.}}
(2006).
Time Ontology in OWL.
\textit{W3C working draft, W3C}.
\end{botherref}
\endbibitem

\bibitem[\protect\citeauthoryear{Pribbenow}{2002}]{pribbenow2002meronymic}
\begin{bchapter}
\bauthor{\bsnm{Pribbenow}, \binits{S.}}
(\byear{2002}).
\bctitle{Meronymic relationships: From classical mereology to complex
  part-whole relations}.
In \bbtitle{The semantics of relationships: An interdisciplinary perspective}
(pp. \bfpage{35}--\blpage{50}).
\bpublisher{Springer}.
\end{bchapter}
\endbibitem

\bibitem[\protect\citeauthoryear{Rodrigues et~al.}{2017}]{rodrigues2017urine}
\begin{bchapter}
\bauthor{\bsnm{Rodrigues}, \binits{F.H.}},
\bauthor{\bsnm{Abel}, \binits{M.}},
\bauthor{\bsnm{Poloni}, \binits{J.A.T.}},
\bauthor{\bsnm{Flores}, \binits{C.D.}} \&
\bauthor{\bsnm{Rotta}, \binits{L.N.}}
(\byear{2017}).
\bctitle{An ontological model for urinary profiles}.
In \bbtitle{2017 IEEE 29th International Conference on Tools with Artificial
  Intelligence (ICTAI)}
(pp. \bfpage{628}--\blpage{635}).
\binstitute{IEEE}.
\end{bchapter}
\endbibitem

\bibitem[\protect\citeauthoryear{Sutcliffe}{2009}]{sutcliffe2009tptp}
\begin{barticle}
\bauthor{\bsnm{Sutcliffe}, \binits{G.}}
(\byear{2009}).
\batitle{The TPTP problem library and associated infrastructure: The FOF and
  CNF parts, v3. 5.0}.
\bjtitle{Journal of Automated Reasoning},
\bvolume{43},
\bfpage{337}--\blpage{362}.
\end{barticle}
\endbibitem

\bibitem[\protect\citeauthoryear{Winston et~al.}{1987}]{winston1987taxonomy}
\begin{barticle}
\bauthor{\bsnm{Winston}, \binits{M.E.}},
\bauthor{\bsnm{Chaffin}, \binits{R.}} \&
\bauthor{\bsnm{Herrmann}, \binits{D.}}
(\byear{1987}).
\batitle{A taxonomy of part-whole relations}.
\bjtitle{Cognitive science},
\bvolume{11}(\bissue{4}),
\bfpage{417}--\blpage{444}.
\end{barticle}
\endbibitem

\bibitem[\protect\citeauthoryear{Zimmerman}{1995}]{zimmerman1995theories}
\begin{barticle}
\bauthor{\bsnm{Zimmerman}, \binits{D.W.}}
(\byear{1995}).
\batitle{Theories of masses and problems of constitution}.
\bjtitle{The philosophical review},
\bvolume{104}(\bissue{1}),
\bfpage{53}--\blpage{110}.
\end{barticle}
\endbibitem

\end{thebibliography}

%

\end{document}